\documentclass[runningheads]{arxiv2608/llncs}
\usepackage{graphicx}
\usepackage[table]{xcolor}
\usepackage{booktabs}
\usepackage{pifont}
\usepackage{amsmath, amssymb, amsfonts}
\usepackage{algorithm}
\usepackage{algorithmic}
\usepackage{multirow}
\usepackage{makecell}
\usepackage{placeins}
\usepackage{changepage}
\usepackage[numbers,sort&compress]{natbib}
\usepackage{adjustbox}
\usepackage[letterpaper, top=3cm, bottom=3cm, left=3.2cm, right=3.2cm]{geometry}
\usepackage{microtype}

\usepackage[breaklinks,colorlinks,citecolor=blue,linkcolor=blue,urlcolor=blue]{hyperref}

\definecolor{myred}{RGB}{190,0,0}
\definecolor{mygreen}{RGB}{0,150,0}
\definecolor{myblue}{RGB}{218,232,246}

\newcommand{\ej}[1]{#1}
\newcommand{\rev}[1]{#1}

\newcommand{\cmark}{\large\color{mygreen}\checkmark}
\newcommand{\xmark}{\large\color{myred}\ding{55}}

\newcommand{\imp}[1]{\,{\textcolor{mygreen}{\scriptsize (#1)}}}
\newcommand{\bad}[1]{\,{\textcolor{myred}{\scriptsize (#1)}}}
\newcommand{\neu}[1]{\,{\textcolor{gray}{\scriptsize (#1)}}}
\newcommand{\vimp}[2]{#1\imp{#2}}
\newcommand{\vbad}[2]{#1\bad{#2}}
\newcommand{\vneu}[2]{#1\neu{#2}}
\newcommand{\bvimp}[2]{\textbf{#1}\imp{#2}}

\newcommand{\samethanks}[1][\value{footnote}]{\footnotemark[#1]}
\makeatletter
\renewcommand\@fnsymbol[1]{\ensuremath{\ifcase#1\or *\or \dagger\or \ddagger\or \mathsection\else?\fi}}
\makeatother

\begin{document}

\title{PatchGate: Narrowing the Verbalization Gap with Intrinsic Object Inventories in Frozen Vision-Language Models}
\titlerunning{PatchGate}

\author{Jihyung Ko\inst{1}\thanks{Equal contribution.\quad\textsuperscript{\ensuremath{\dagger}}\,Corresponding authors: \texttt{shjo.april@gmail.com, kyskim@snu.ac.kr}} \and
Eunji Jung\inst{1}\samethanks \and
Hyeongsub Kim\inst{1, 2} \and
Ziseok Lee\inst{1} \and \\
Jae Won Cho\inst{3} \and
Sanghyun Jo\inst{1,4}\samethanks[2] \and
Kyungsu Kim\inst{1}\samethanks[2]}
\authorrunning{J. Ko et al.}

\institute{$^{1}$Seoul National University, Korea \quad $^{2}$LG CNS, Korea \quad $^{3}$Konkuk University, Korea \quad $^{4}$OGQ, Korea}

\maketitle

\begin{abstract}
{Reliable image captioning in Vision-Language Models (VLMs) requires captions to be both precise and complete, avoiding unsupported object mentions while covering visible objects. Existing training-free methods primarily address the former requirement, suppressing unsupported object words by intervening on model-predicted mentions during generation. Because they operate only on objects the model is already likely to mention, visible objects omitted from the output remain difficult to recover. We propose \textbf{PatchGate}, a training-free framework that extracts prompt-free object evidence intrinsic to a frozen VLM before generation and uses it to narrow the gap between an intrinsic object set and final object mentions. In the first stage, Visual Evidence eXtraction (VEX) reads patch-level lexical evidence from the latter half of LM decoder layers and constructs an image-conditioned object set without any task prompt. In the second stage, Visual-Evidence Inclusion-Exclusion Decoding (VIED) uses this object evidence to calibrate decoding logits, promoting evidence-supported but under-verbalized objects and suppressing weakly supported but over-verbalized objects. On AMBER, PatchGate improves both sides of object-level reliability, increasing visible-object coverage from \(49.4\) to \(56.0\) (\(+13.4\%\)) and reducing object hallucination by lowering CHAIR from \(7.5\) to \(6.6\) (\(-12.0\%\)), without external detectors or fine-tuning and with one extra forward pass.
}
% \keywords{Vision-Language Models \and Object Hallucination \and Object Omission \and Training-free Decoding}
\end{abstract}

\begin{figure}[t]
    \centering
    \includegraphics[width=1.0\linewidth]{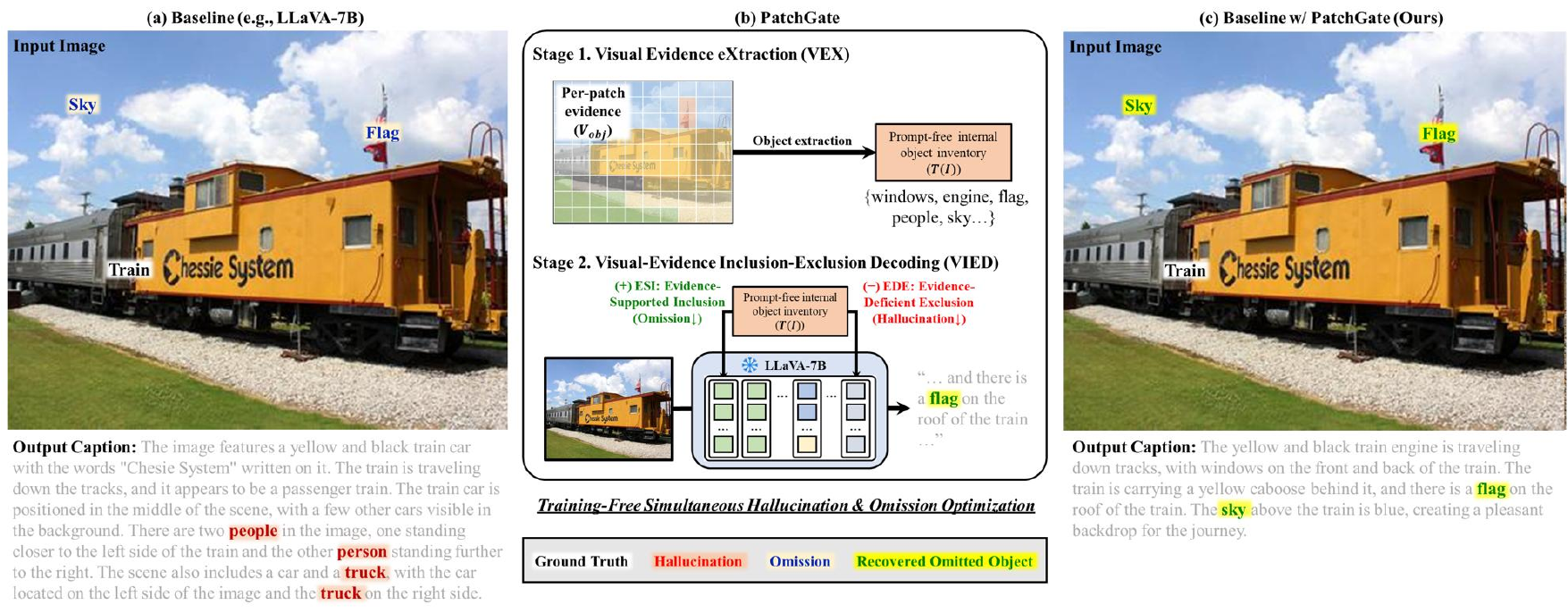}
    \caption{{\textbf{Two complementary captioning failures and PatchGate's simultaneous mitigation.} Reliable captions should be precise and complete, yet generated captions can mention unsupported objects (red: \emph{people}, \emph{truck}) while omitting visible objects (blue: \emph{sky}, \emph{flag}). PatchGate uses prompt-free internal object evidence to calibrate decoding, suppressing unsupported mentions and recovering omitted visible objects (yellow: \emph{sky}, \emph{flag}).}}
    \label{fig:baseline_vs_ours}
\end{figure}

\section{Introduction} \label{sec:intro}
{Modern Vision-Language Models (VLMs)~\cite{liu2024llavaimproved, Qwen2.5-VL} have become a common interface for image understanding, generating language descriptions, answers, and explanations grounded in visual inputs. As these models are increasingly used to describe visual content, their reliability depends not only on linguistic fluency but also on whether the generated text remains faithful to the image.}

{For image captioning, faithful outputs should be both precise and complete, avoiding unsupported object mentions while covering visible objects.
However, generated captions often violate these requirements in two complementary ways: they can mention unsupported objects or omit visible objects needed for a complete description, commonly referred to as object hallucination and object omission, respectively (see Fig.~\ref{fig:baseline_vs_ours}).}

{Existing training-free methods seek more faithful captions by exploiting generation-time signals, including attention maps~\cite{huang2024opera} and contrastive decoding distributions~\cite{leng2024vcd}. However, these signals mainly act on object words the model is already likely to generate, making them effective at suppressing unsupported object mentions but limited at recovering visible objects that were omitted. Other methods~\cite{yin2024woodpecker, zhao2025marine} leverage external detectors~\cite{liu2024groundingdinotiny, detr} to supply missing object cues, but these cues come from external visual modules rather than the target VLM's own internal evidence, introducing additional modules and dependence on detector coverage and accuracy.}

{We examine whether the frozen VLM contains object evidence before generation, and how this evidence differs from the objects ultimately mentioned in the caption.
Building on logit-lens analyses in language models~\cite{geva2022transformer, belrose2023eliciting} and their extension to VLM patch representations~\cite{neo2025towards}, we read lexical evidence from internal patch hidden states without any task prompt.
This comparison reveals an evidence-verbalization mismatch: evidence-supported visible objects can remain unmentioned, leading to object omissions, while weakly supported object words can still be mentioned, leading to object hallucinations. This suggests a shared basis for recovering under-verbalized visible objects and suppressing weakly supported object mentions.}

{Based on this observation, we propose PatchGate, a two-stage training-free framework that uses the frozen VLM's internal patch representations to narrow the evidence-verbalization mismatch.
The first stage, Visual Evidence eXtraction (VEX), reads patch-level lexical evidence from the latter half of the LM decoder layers, scores each object candidate by its strongest normalized patch evidence, and constructs an intrinsic object inventory before generation, defined as a prompt-free, image-conditioned set of visually supported objects. The second stage, Visual-Evidence Inclusion-Exclusion Decoding (VIED), uses this inventory to calibrate decoding logits, jointly promoting evidence-supported but under-verbalized objects and suppressing weakly supported object mentions (\rev{Fig.~\ref{fig:baseline_vs_ours}}). PatchGate thereby improves both caption precision and completeness without external detectors or fine-tuning.}

{Our contributions are summarized as follows:}

\begin{itemize}
    \item {We characterize an object-level evidence-verbalization mismatch in frozen VLMs by comparing prompt-free internal object evidence with final object mentions: visual-evidence deficits among mentioned object words indicate unsupported mentions, while strong evidence among unmentioned objects indicates omitted visible objects.}

    \item {We introduce \textbf{PatchGate}, a training-free framework that narrows this mismatch by extracting an intrinsic object inventory before generation and using it to calibrate decoding logits for more precise and complete captions.}

    \item {PatchGate improves object-level reliability without external detectors or fine-tuning, increasing AMBER Cover by \(13.4\%\), lowering CHAIR by \(12.0\%\), and boosting POPE Precision/Recall by $2.3\%$/$20.6\%$ with consistent gains across backbones.}
\end{itemize}

\begin{table}[t]
    \centering\small
    \setlength{\tabcolsep}{2pt}
    \renewcommand{\arraystretch}{1.15}
    \caption{{Conceptual comparison of PatchGate (Ours) with representative training-free hallucination mitigation approaches: OPERA~\cite{huang2024opera}, VCD~\cite{leng2024vcd}, ProjectAway~\cite{jiang2025projectaway}, MARINE~\cite{zhao2025marine}, SHIELD~\cite{huang2026shield}, PND~\cite{jiang2026pnd} and ILVAD~\cite{Xie2026ilvad}.}}
    \label{tab:conceptual_comparision}
    \resizebox{\linewidth}{!}{
    \begin{tabular}{l|cccccccc}
    \toprule
    \rowcolor{gray!15}
    Properties
     & \shortstack{[CVPR'24]\\OPERA}
     & \shortstack{[CVPR'24]\\VCD}
     & \shortstack{[ICLR'25]\\ProjectAway}
     & \shortstack{[ICML'25]\\MARINE}
     & \shortstack{[ICLR'26]\\SHIELD}
     & \shortstack{[CVPR'26]\\PND}
     & \shortstack{[ICML'26]\\ILVAD}
     & \shortstack{\textbf{PatchGate}\\\textbf{(Ours)}} \\
    \midrule
    (a)\ Reduces object hallucinations for precision
     & \cmark & \cmark & \cmark & \cmark & \cmark & \cmark & \cmark & \cmark \\
    (b)\ Reduces object omissions for completeness
     & \xmark & \xmark & \xmark & \xmark & \xmark & \cmark & \xmark & \cmark \\
    (c)\ Reads internal patch-level object evidence
     & \xmark & \xmark & \cmark & \xmark & \xmark & \xmark & \xmark & \cmark \\
    (d)\ Builds a prompt-free object set before generation
     & \xmark & \xmark & \xmark & \cmark & \xmark & \xmark & \xmark & \cmark \\
    (e)\ Corrects visual evidence-mention mismatch
     & \xmark & \xmark & \xmark & \xmark & \xmark & \xmark & \xmark & \cmark \\
    (f)\ Requires no external visual module (\emph{e.g.}, tagger)
     & \cmark & \cmark & \cmark & \xmark & \cmark & \xmark & \cmark & \cmark \\
    \bottomrule
    \end{tabular}
    }
\end{table}

\section{Related Work} \label{sec:related_work}
\subsection{Hallucination Mitigation in VLMs} \label{subsec:related_work_hallucination_method}

{Reliable Vision-Language Models (VLMs) require captions to be both precise and complete: they should avoid object hallucinations, where captions mention objects unsupported by the image, while covering visible objects without omissions.
Prior work has sought to improve this reliability through two broad approaches: training-based and training-free. Training-based methods~\cite{xieetal2024vdpo,Yang2025opadpo} reduce hallucination through additional supervision or preference optimization, but require costly model updates and extra training data.
Accordingly, training-free methods have gained traction by keeping the VLM fixed and modifying inference-time signals, offering a practical plug-and-play alternative.}

{Representative training-free methods mitigate object hallucination through contrastive decoding~\cite{leng2024vcd,chen2024halc,huo2024sid,wang2024icd}, attention intervention~\cite{huang2024opera,liu2024pai,an2025agla,jiang2025devils,Yu2026owl,Xie2026ilvad}, visual encoder correction~\cite{huang2026shield}, or external object guidance~\cite{zhao2025marine,yin2024woodpecker}.
Because these methods mainly act on objects the model is already likely to mention, they are effective at reducing hallucination but limited in recovering visible objects that never surface in the generated text.
Recent work~\cite{jiang2026pnd} also considers both hallucination and omission, but because it corrects object candidates only as they arise during generation, visible objects not covered by the caption remain difficult to identify and recover.
PatchGate instead uses a pre-generation intrinsic object set to form a signed evidence-verbalization gap, whose two directions drive evidence-supported inclusion for omission recovery and evidence-deficient exclusion for hallucination suppression, jointly improving caption completeness and precision (see Tab.~\ref{tab:conceptual_comparision}).}

\subsection{Internal Visual Evidence} \label{subsec:related_work_internal_evidence}

{Internal representations provide a useful window into what a frozen model encodes before it verbalizes.
In language models, logit-lens and tuned-lens analyses show that intermediate hidden states can expose lexical predictions before the final decoding layer~\cite{geva2022transformer,belrose2023eliciting}.
Recent VLM studies~\cite{neo2025towards} extend this view to visual tokens, showing that object information can be localized in patch representations and becomes increasingly readable in the vocabulary space across layers. Building on these findings, PatchGate studies the object-level gap between pre-generation visual evidence and final object mentions.}

{Some studies use internal visual evidence for hallucination analysis, detection, grounding, or probing~\cite{phukan2025contextuallens, kogilathota2026halp}, while mitigation methods use internal representations for editing hallucination-related directions or steering generation with intermediate-layer logits and token-ranking signals~\cite{jiang2025projectaway, wang2025deco, li2025vista}.
These works show that internal states can expose object-level visual support, but such support is typically used as an absolute grounding signal for an object.
Absolute support alone does not determine how the caption should be corrected: a supported object may still need to be promoted if it is omitted, while a weakly supported object should matter only when it is over-verbalized.
PatchGate therefore compares pre-generation object evidence with object verbalization in the same object coordinate, yielding a signed evidence-verbalization residual whose two directions drive evidence-supported inclusion and evidence-deficient exclusion.}

\begin{figure}[t]
    \centering
    \includegraphics[width=1.0\linewidth]{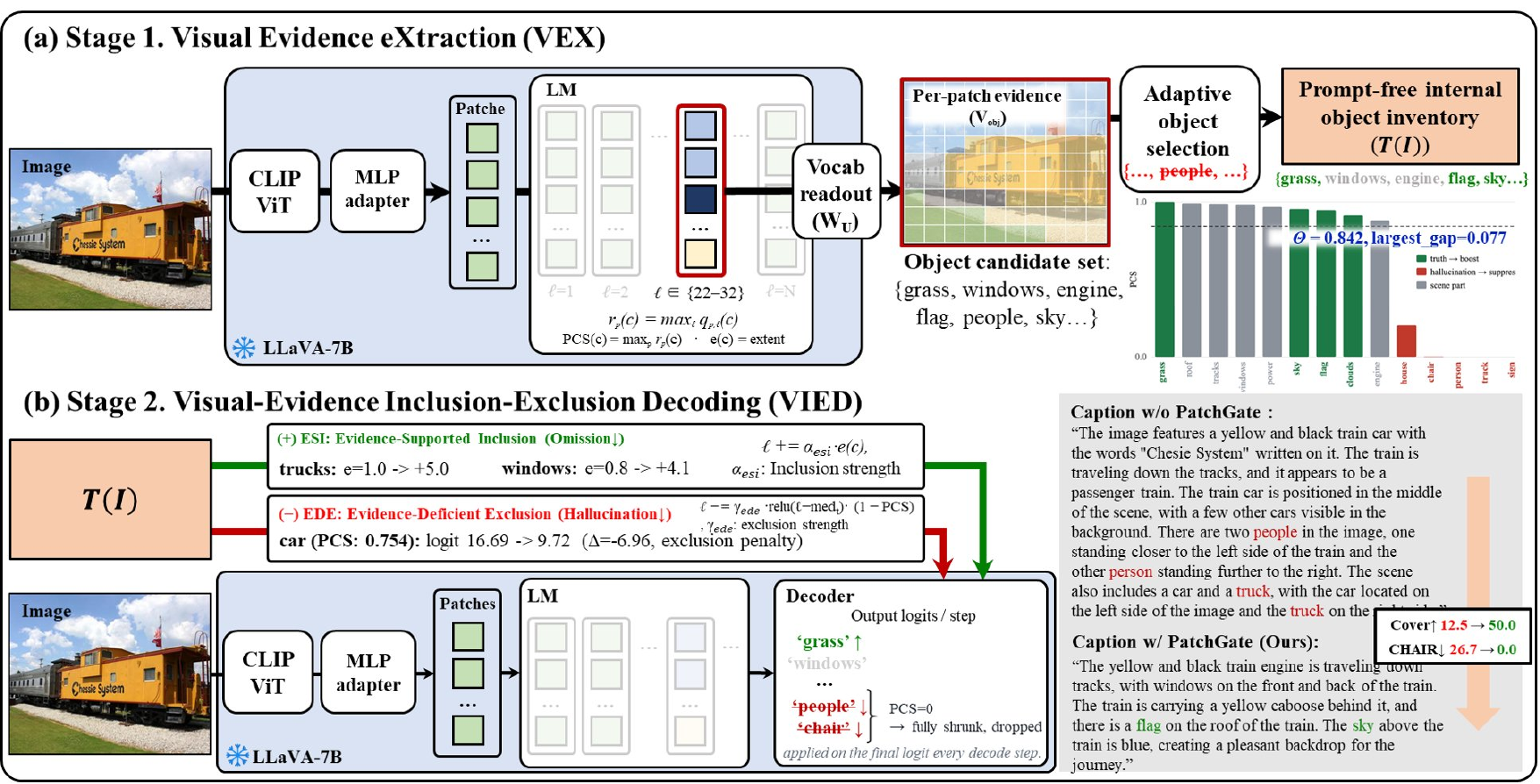}
    \caption{\textbf{Overview of PatchGate.}
    (a) VEX reads patch-level lexical evidence from the later LM decoder layers, scores each object candidate by its strongest normalized patch evidence, and selects a prompt-free object set \(\mathcal{T}(I)\) with an adaptive largest-gap cutoff.
    (b) VIED calibrates decoding logits with \(\mathcal{T}(I)\): inclusion promotes evidence-supported but under-verbalized objects, while exclusion suppresses over-committed object words with weak visual evidence.}
    \label{fig:method}
\end{figure}

\section{Method} \label{sec:method}
{PatchGate is a two-stage training-free framework that aligns a frozen VLM's internal object evidence with its final object verbalization to mitigate object hallucinations and omissions simultaneously.
Visual Evidence eXtraction (VEX; Sec.~\ref{subsec:vex}) first extracts a prompt-free, image-conditioned object inventory \(\mathcal{T}(I)\) from internal patch representations before caption generation.
Visual-Evidence Inclusion-Exclusion Decoding (VIED; Sec.~\ref{subsec:vied}) then uses this inventory to calibrate decoding logits in two complementary directions, promoting visually supported but under-verbalized objects and suppressing over-verbalized object words with weak visual evidence.
The overview of PatchGate is shown in Fig.~\ref{fig:method}, with implementation details provided in Appendix~\ref{app:implementation_detail}.}

\subsection{Visual Evidence eXtraction (VEX)} \label{subsec:vex}
{VEX constructs an intrinsic object inventory \(\mathcal{T}(I)\), a prompt-free, image-conditioned set of visually supported objects, through three steps: patch-to-vocabulary readout, object evidence scoring, and adaptive object selection.}

\paragraph{Step 1: Patch-to-vocabulary readout.}

{In this step, VEX projects each patch hidden state into the model's output vocabulary using the logit lens~\cite{geva2022transformer} and collects image-specific object candidates from top-1 lexical readouts.
Given an image \(I\), let \(h_p^{(\ell)}\) be the hidden state of patch \(p\in\{1,\dots,N\}\) at LM decoder layer \(\ell\).
Following prior observations that later-layer predictions are more interpretable and better aligned with final predictions~\cite{belrose2023eliciting}, we read from a later layer range \(\mathcal{L}\).
For our default backbone, LLaVA-v1.5-7B~\cite{liu2024llavaimproved}, the decoder layers are indexed from \(1\) to \(32\), and we use \(\mathcal{L}=\{22,\dots,32\}\) with \(N=576\) visual patches.
(Appendix~\ref{app:ablation_study} ablates this layer range.)
For each selected layer and patch, we define the probability assigned to vocabulary token \(v\) as}
\begin{equation}
q^{(\ell)}_p(v)
=
\mathrm{softmax}\!\left(
W_U\,\mathrm{RMSNorm}(h_p^{(\ell)})
\right)[v],
\quad
v\in\mathcal{V}.
\label{eq:lens}
\end{equation}

{Here, \(\mathcal{V}\) is the full language vocabulary, and \(W_U\) is the frozen LM unembedding matrix applied after RMSNorm~\cite{zhang2019rmsnorm}.
Let \(\mathcal{V}_{\mathrm{obj}}\) denote the WordNet physical-object vocabulary~\cite{miller1995wordnet}.
We retain top-1 readouts that belong to \(\mathcal{V}_{\mathrm{obj}}\), forming the image-specific candidate set}
\begin{equation}
\mathcal{C}(I)
=
\left\{
\arg\max_{v\in\mathcal{V}} q^{(\ell)}_p(v)
:
\ell\in\mathcal{L},\ p\in\{1,\dots,N\}
\right\}
\cap
\mathcal{V}_{\mathrm{obj}}.
\label{eq:candidate_set}
\end{equation}

\paragraph{Step 2: Object evidence scoring.}
{VEX assigns each object word a scalar visual evidence score for later evidence-verbalization correction.
For any object word \(c\in\mathcal{V}_{\mathrm{obj}}\), we define its patch-confidence score as the strongest evidence over the selected layers and patches:}
\begin{equation}
\mathrm{PCS}(c)
=
\max_{p\in\{1,\dots,N\}}
\max_{\ell\in\mathcal{L}}
q^{(\ell)}_p(c).
\label{eq:pcs}
\end{equation}
{\(\mathrm{PCS}(c)\) emphasizes localized high-confidence evidence over patch-readout frequency, helping preserve small-region candidates while providing the shared evidence score for inventory selection and VIED calibration (Sec.~\ref{subsec:vied}).
Because \(\mathrm{PCS}\) maximizes over both patches and layers, no single patch or layer is selected: an object receives high evidence if any late layer reads it strongly in any image region.}

\paragraph{Step 3: Adaptive object selection.}
{The final object inventory is selected with an image-adaptive cutoff.
Instead of using a fixed top-\(k\), we set \(\theta_{\mathrm{TTD}}(I)\) as the midpoint of the largest gap between sorted raw \(\mathrm{PCS}\) values, following TTD~\cite{jo2024ttd}.}
\begin{equation}
\mathcal{T}(I)
=
\{c\in\mathcal{C}(I):\mathrm{PCS}(c)>\max(\theta_{\mathrm{TTD}}(I), \tau_{\mathrm{floor}})\}.
\label{eq:inventory}
\end{equation}
{The floor \(\tau_{\mathrm{floor}}\) filters weakly supported candidates when all scores are low, while sorted-tail exclusion prevents outlier-driven thresholds.
The exact cutoff procedure is provided in Appendix~\ref{app:methodological_details}.}
{After these steps, VEX yields \(\mathcal{T}(I)\) as an intrinsic, prompt-free object inventory extracted before caption generation. It reflects object evidence available inside the frozen VLM and serves as the reference for VIED to compare internal support with final object verbalization.}

\begin{table}[t]
    \centering
    \caption{Performance comparison on AMBER generative and discriminative tasks~\cite{wang2023amber}.
    For Cover, CHAIR, Precision (P.), and Recall (R.), colored values show absolute changes from the LLaVA-v1.5-7B baseline~\cite{liu2024llavaimproved} (green: improvement, red: degradation). The best results are in \textbf{bold}.}
    \label{tab:main_amber}
    \resizebox{\linewidth}{!}{
    \begin{tabular}{lcccccccc}
    \toprule
    \multirow{2}{*}{\textbf{Method}}
    & \multicolumn{4}{c}{\textbf{Generative Task (Captioning)}}
    & \multicolumn{4}{c}{\textbf{Discriminative Task (QA)}} \\
    \cmidrule(lr){2-5} \cmidrule(lr){6-9}
    & \textbf{Cover} $\uparrow$
    & \textbf{CHAIR} $\downarrow$
    & \textbf{Hal} $\downarrow$
    & \textbf{Cog} $\downarrow$
    & \textbf{Acc.} $\uparrow$
    & \textbf{P.} $\uparrow$
    & \textbf{R.} $\uparrow$
    & \textbf{F1} $\uparrow$ \\
    \midrule
    {[CVPR'24] LLaVA-v1.5-7B}
    & 49.4 & 7.5 & 31.4 & 3.6
    & 72.0 & 92.5 & 62.9 & 74.9 \\
    { + [CVPR'24] OPERA}
    & \vbad{48.8}{-0.6}
    & \vimp{6.6}{-0.9}
    & 28.3 & 2.9
    & \textbf{76.4} & \vimp{92.7}{+0.2} & \vimp{69.9}{+7.0} & \textbf{79.7} \\
    { + [CVPR'24] VCD}
    & \vimp{51.7}{+2.3}
    & \vbad{8.9}{+1.4}
    & 39.7 & 4.4
    & 71.1 & \vbad{91.4}{-1.1} & \vbad{62.2}{-0.7} & 74.0 \\
    { + [CVPR'25] Devils-in-Mid.}
    & \vimp{50.3}{+0.9}
    & \bvimp{3.3}{-4.2}
    & 20.0 & \textbf{1.2}
    & 50.5 & \vimp{94.5}{+2.0} & \vbad{26.8}{-36.1} & 41.8 \\
    { + [ICML'25] MARINE}
    & \vimp{49.9}{+0.5}
    & \vimp{6.0}{-1.5}
    & 27.3 & 2.9
    & 72.4 & \bvimp{94.8}{+2.3} & \vbad{61.7}{-1.2} & 74.7 \\
    { + [ICLR'25] ProjectAway}
    & \vneu{49.4}{+0.0}
    & \vbad{9.2}{+1.7}
    & 37.6 & 5.0
    & 70.6 & \vneu{92.5}{+0.0} & \vbad{60.6}{-2.3} & 73.2 \\
    { + [ICML'26] ILVAD}
    & \vbad{44.9}{-4.5}
    & \vimp{4.1}{-3.4}
    & \textbf{19.0} & 1.3
    & 75.4 & \vbad{90.5}{-2.0} & \bvimp{70.3}{+7.4} & 79.1 \\
    { + [ICLR'26] SHIELD}
    & \vimp{54.5}{+5.1}
    & \vbad{8.5}{+1.0}
    & 42.5 & 4.0
    & 64.5 & \vimp{93.5}{+1.0} & \vbad{50.0}{-12.9} & 65.2 \\
    \rowcolor{myblue}
    \textbf{ + PatchGate (Ours)}
    & \bvimp{56.0}{+6.6}
    & \vimp{6.6}{-0.9}
    & 34.3 & 3.0
    & 74.5 & \vimp{93.8}{+1.3} & \vimp{66.0}{+3.1} & 77.4 \\
    \bottomrule
    \end{tabular}
    }
\end{table}

\subsection{Visual-evidence Inclusion-Exclusion Decoding (VIED)} \label{subsec:vied}

{VIED uses the VEX-extracted object inventory \(\mathcal{T}(I)\) (Sec.~\ref{subsec:vex}) to calibrate decoding logits according to the evidence-verbalization mismatch.
This mismatch has two complementary directions: visually supported objects can be under-verbalized and omitted from the caption, while weakly supported object words can be over-verbalized and appear as hallucinations.
Accordingly, VIED applies two logit-space operations during decoding: Evidence-Supported Inclusion (ESI) promotes objects supported by \(\mathcal{T}(I)\), and Evidence-Deficient Exclusion (EDE) suppresses object words with high decoding logits despite weak visual evidence.
Both operations are applied without modifying the prompt, querying an external model, or updating model parameters.}

\paragraph{(+) ESI: Evidence-Supported Inclusion.}
{Let \(g_t[v]\) denote the base logit of token \(v\) at decoding step \(t\).
ESI promotes objects in \(\mathcal{T}(I)\) that are visually supported but may remain under-verbalized during generation.
For each object \(c\in\mathcal{T}(I)\), we use an extent score \(e(c)\in[0,1]\), defined as the inventory-normalized fraction of patches whose strongest object evidence is assigned to \(c\):}
\begin{equation}
\begin{aligned}
\tilde{e}(c)
&=
\frac{1}{N}
\sum_{p=1}^{N}
\mathbf{1}\!\left[
c=
\operatorname*{arg\,max}_{c'\in\mathcal{V}_{\mathrm{obj}}}
\max_{\ell\in\mathcal{L}}
q_p^{(\ell)}(c')
\right],
\\
e(c)
&=
\frac{\tilde{e}(c)}
{\max_{c'\in\mathcal{T}(I)}\tilde{e}(c')}.
\end{aligned}
\label{eq:esi_bonus}
\end{equation}
{The inner maximum collapses the selected layer range before patch assignment, so each patch casts one object vote and the denominator of \(\tilde{e}(c)\) is \(N\), not \(N|\mathcal{L}|\).
This extent score determines the ESI bonus, which is applied at each decoding step until the corresponding object is generated.}

\paragraph{(-) EDE: Evidence-Deficient Exclusion.}
{EDE suppresses object words that are strongly favored by the decoder despite weak visual evidence, since such words can surface as object hallucinations.
Rather than applying a hard veto to objects outside \(\mathcal{T}(I)\), EDE applies a visual-evidence-gated exclusion signal over the object vocabulary.
At decoding step \(t\), the median logit \(\mathrm{med}_t=\mathrm{median}_{v\in\mathcal{V}}g_t[v]\) serves as a neutral reference level.
For each object word \(c\in\mathcal{V}_{\mathrm{obj}}\), \(\mathrm{relu}(g_t[c]-\mathrm{med}_t)\) measures verbalization excess above this neutral level, while \(1-\mathrm{PCS}(c)\) measures visual deficit.
The exclusion signal multiplies these two factors:}
\begin{equation}
\Delta_t^{-}(c)
=
\underbrace{\mathrm{relu}\!\left(g_t[c]-\mathrm{med}_t\right)}_{\text{verbalization excess}}
\underbrace{\big(1-\mathrm{PCS}(c)\big)}_{\text{visual deficit}},
\quad c\in\mathcal{V}_{\mathrm{obj}}.
\label{eq:ede_penalty}
\end{equation}
{Because \(\Delta_t^{-}(c)\) depends on continuous visual evidence, object words with higher \(\mathrm{PCS}\) receive weaker suppression even if they fall outside the final inventory.
Thus, EDE reduces reliance on the hard inventory boundary while targeting above-neutral object logits with weak visual evidence.}

\paragraph{Unified logit update.}
{ESI and EDE form a single logit edit at each decoding step.
Let \(\mathcal{M}_{<t}\subseteq\mathcal{V}_{\mathrm{obj}}\) denote the object words already generated before step \(t\).
For every token \(v\), VIED applies \(\hat{g}_t[v]=g_t[v]+\eta_t(v)\), where}
\begin{equation}
\eta_t(v)
=
\left\{
\begin{array}{l @{\mspace{-40mu}} r@{}}
\alpha_{\mathrm{esi}}e(v)\mathbf{1}[v\notin\mathcal{M}_{<t}] -\gamma_{\mathrm{ede}}\Delta_t^{-}(v), & v\mkern-3mu\in\mkern-3mu\mathcal{T}(I),\\[2pt]
-\gamma_{\mathrm{ede}}\Delta_t^{-}(v), & v\mkern-3mu\in\mkern-3mu\mathcal{V}_{\mathrm{obj}}\setminus\mathcal{T}(I),\\[2pt]
0, & v\mkern-3mu\notin\mkern-3mu\mathcal{V}_{\mathrm{obj}}.
\end{array}
\right.
\label{eq:vied_unified}
\end{equation}
{Here, \(\alpha_{\mathrm{esi}}\) and \(\gamma_{\mathrm{ede}}\) control the inclusion and exclusion strengths, respectively.
Thus, ESI promotes only not-yet-generated inventory objects, EDE applies a visual-evidence-gated penalty to object words, and non-object tokens are unchanged. Together, the edits align pre-generation object evidence with step-wise object verbalization.}

\section{Experiments} \label{sec:experiments}

\subsection{Experimental Setup} \label{subsec:experimental_setup}
\paragraph{Models and baselines.}
{We evaluate PatchGate on the frozen LLaVA-v1.5-7B backbone~\cite{liu2024llavaimproved}.
Under the same benchmark protocols, we compare against the original LLaVA baseline and representative training-free methods, including OPERA~\cite{huang2024opera}, VCD~\cite{leng2024vcd}, Devils-in-Mid.~\cite{jiang2025devils}, MARINE~\cite{zhao2025marine}, ProjectAway~\cite{jiang2025projectaway}, ILVAD~\cite{Xie2026ilvad}, and SHIELD~\cite{huang2026shield}.}

{For ablations, we further test external tag extraction with GroundingDINO~\cite{liu2024groundingdinotiny} and RAM++~\cite{huang2025ram++}, and backbone robustness with \rev{LLaVA-v1.5-13B~\cite{liu2024llavaimproved},} Qwen2.5-VL~\cite{Qwen2.5-VL}\rev{,} and InstructBLIP~\cite{dai2023instructblip}.}

\paragraph{Benchmarks.}
{We evaluate PatchGate on AMBER~\cite{wang2023amber} and POPE~\cite{li2023pope}, which cover complementary object-level reliability settings.
AMBER includes free-form captioning and discriminative object-existence question-answering (QA).
In captioning, no object query is given, so generated captions are evaluated for visible-object coverage and unsupported object mentions.
In AMBER QA and POPE, the queried object is given, and the model answers Yes/No for binary object-existence.}

\paragraph{Metrics.}
{We use task-specific metrics according to each benchmark output format.
For AMBER's generative captioning task, we follow the official protocol and report Cover, CHAIR, Hal, and Cog using the provided object annotations.
We treat Cover and CHAIR~\cite{rohrbach2018chair} as the primary generative metrics because they directly track our two target axes: covering visible objects and avoiding unsupported object mentions.
Hal and Cog are complementary diagnostics: Hal is a caption-level any-hallucination indicator, while Cog depends on AMBER's predefined hallucinatory target object set.
For AMBER's discriminative task and POPE, we report Accuracy (Acc.), Precision (P.), Recall (R.), and F1 for binary object-existence answers.
Detailed metric definitions are provided in Appendix~\ref{app:experimental_details}.}

\begin{table}[t]
    \centering
    \caption{Average performance on POPE across random, popular, and adversarial splits. The best results are in \textbf{bold}.}
    \label{tab:main_pope_avg}
    \adjustbox{max width=\linewidth}{
    \begin{tabular}{lcccc}
    \toprule
    \textbf{Method}
    & \textbf{Acc.} $\uparrow$
    & \textbf{P.} $\uparrow$
    & \textbf{R.} $\uparrow$
    & \textbf{F1} $\uparrow$ \\
    \midrule
    {[CVPR'24] LLaVA-v1.5-7B}
    & 82.0 & \rev{88.6} & 73.7 & 80.4 \\
    { + [CVPR'24] OPERA}
    & 85.7 & 86.7 & 85.2 & 85.9 \\
    { + [CVPR'24] VCD}
    & 83.0 & \rev{88.0} & 76.8 & 81.9 \\
    { + [CVPR'25] Devils-in-Mid.}
    & 55.6 & 54.3 & 70.1 & 61.2 \\
    { + [ICML'25] MARINE}
    & 84.4 & 89.4 & 78.0 & 83.3 \\
    { + [ICLR'25] ProjectAway}
    & 84.3 & \rev{81.9} & \textbf{89.2} & \rev{85.3} \\
    { + [ICLR'26] SHIELD}
    & 86.4 & \rev{85.2} & 88.6 & 86.7 \\
    { + [ICML'26] ILVAD}
    & 85.6 & \rev{85.2} & 86.6 & 85.7 \\
    \rowcolor{myblue}
    \textbf{ + PatchGate (Ours)}
    & \textbf{89.8} & \textbf{\rev{90.5}} & 88.9 & \textbf{\rev{89.6}}\\
    \bottomrule
    \end{tabular}
    }
\end{table}

\begin{figure}[t]
    \centering
    \includegraphics[width=0.95\linewidth]{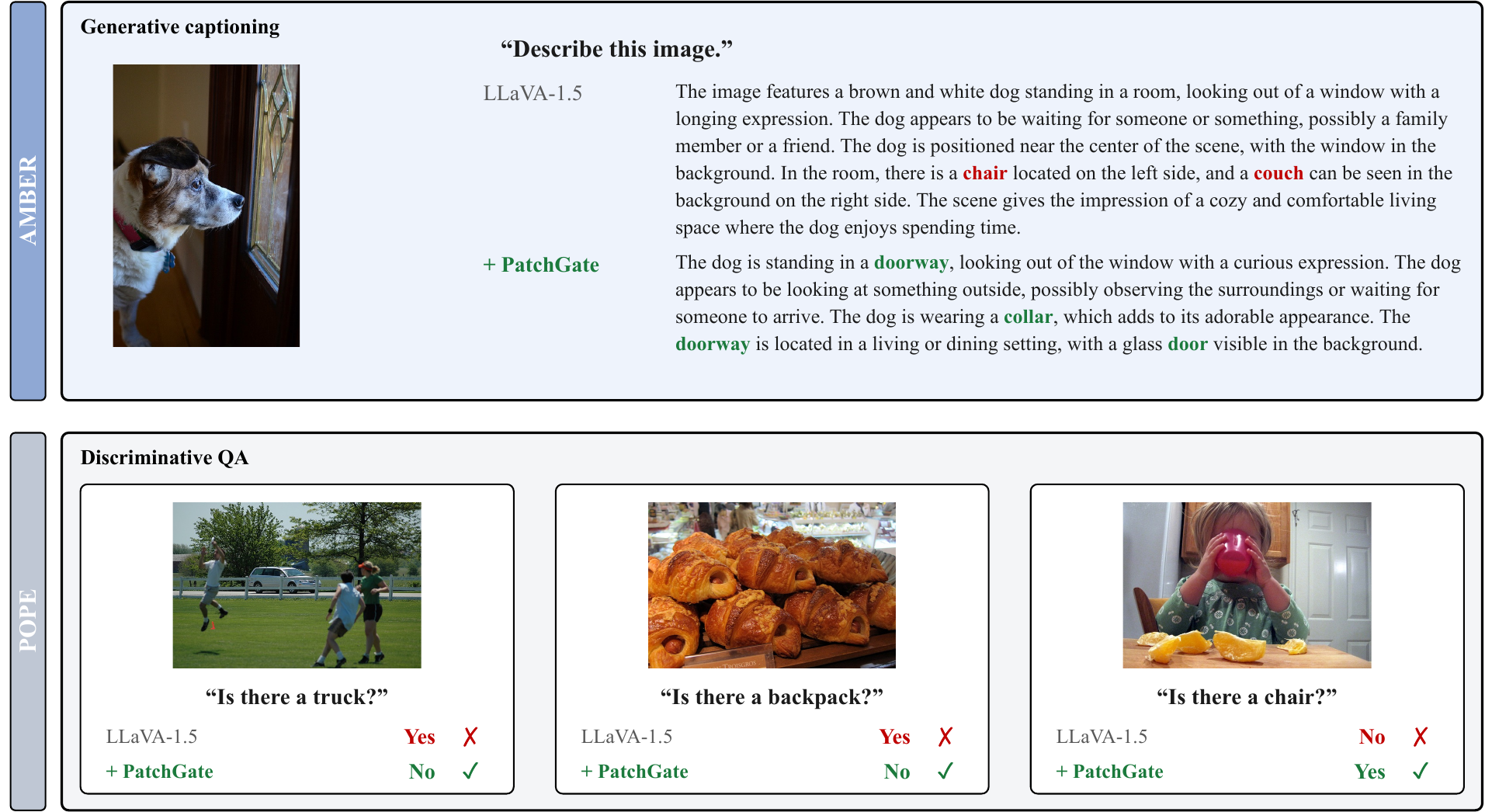}
    \caption{\textbf{Qualitative results on AMBER~\cite{wang2023amber} and POPE~\cite{li2023pope}.} PatchGate recovers omitted visible objects and enables correct predictions (green), while suppressing unsupported object mentions and incorrect answers (red).}
    \label{fig:qualitative_results}
\end{figure}

\subsection{Quantitative Results} \label{subsec:quantitative_results}

\paragraph{AMBER.}
{Tab.~\ref{tab:main_amber} shows that PatchGate is the only method that improves all four highlighted axes over the LLaVA-v1.5-7B baseline: Cover, CHAIR, P., and R.
Prior methods often trade off one axis for another: SHIELD improves Cover and P. but worsens CHAIR and R.; OPERA improves CHAIR but reduces Cover; and ILVAD improves CHAIR and R. while degrading Cover and P. PatchGate instead improves all four axes, increasing Cover from \(49.4\) to \(56.0\) (\(+13.4\%\)), reducing CHAIR from \(7.5\) to \(6.6\) (\(-12.0\%\)), and improving P. and R. to \(93.8\) and \(66.0\), respectively.
This supports our main claim that PatchGate improves caption completeness and precision more consistently across both generative captioning and discriminative QA (See Appendix~\ref{app:quantitative_results} for additional baseline evaluations and extended quantitative results).}

{Hal and Cog are not uniformly improved, so PatchGate does not dominate every AMBER diagnostic.
Hal depends on caption-level any-hallucination decisions, while Cog depends on AMBER's predefined target object set. We therefore interpret them as complementary diagnostics and focus the main generative comparison on Cover and CHAIR.}

\paragraph{POPE.}
{Tab.~\ref{tab:main_pope_avg} shows that PatchGate achieves the best average Acc., P., and F1 on POPE.
Compared with the LLaVA-v1.5-7B baseline, PatchGate improves Acc. from \(82.0\) to \(89.8\) (\(+9.5\%\)), P. from \rev{\(88.5\)} to \rev{\(90.5\)} (\(+2.3\%\)), and F1 from \(80.4\) to \rev{\(89.6\)} (\rev{\(+11.4\%\)}).
It also achieves second-best on R. from \(73.7\) to \(88.9\), only \(0.3\) points below the best result from ProjectAway, while achieving stronger overall balance across the other metrics.
This suggests that PatchGate improves binary object-existence reliability without merely biasing the model toward either "Yes" or "No".
Split-wise POPE results are provided in Appendix~\ref{app:quantitative_results}.}

\subsection{Qualitative Results} \label{subsec:qualitative_results}
{Qualitative examples further illustrate PatchGate's two-sided correction behavior (see Fig.~\ref{fig:qualitative_results}).
In AMBER generative captioning, PatchGate suppresses hallucinated object mentions while recovering omitted visible objects, improving both caption precision and completeness.
In POPE discriminative QA, it corrects incorrect "Yes" answers to absent objects and incorrect "No" answers to present objects, rather than introducing a one-sided answer bias.
This matches the trends in Tabs.~\ref{tab:main_amber} and~\ref{tab:main_pope_avg}, where PatchGate improves omission-related metrics, Cover and R., together with hallucination-related metrics, CHAIR and P.}

\subsection{Ablation Study} \label{subsec:ablation_study}
\begin{table}[t]
    \centering
    \caption{{Diagnostic ability of pre-decoding VEX evidence (Sec.~\ref{subsec:vex}) for hallucination and omission errors. Random baselines: AUROC \(=0.5\), AUPRC \(=\) positive rate.}}
    \label{tab:ab_mismatch}
    \adjustbox{max width=\linewidth}{
    \begin{tabular}{llccc}
    \toprule
    \textbf{Condition}
    & \textbf{Error target}
    & \textbf{Ranker}
    & \textbf{AUROC} $\uparrow$
    & \textbf{AUPRC} $\uparrow$ \\
    \midrule
    Mentioned
    & Hallucination
    & Random
    & 0.500
    & 0.163 \\
    \rowcolor{myblue}
    Mentioned
    & Hallucination
    & \textbf{\(1-\mathrm{PCS}\)}
    & \textbf{0.883}
    & \textbf{0.707} \\
    \midrule
    Unmentioned
    & Omission
    & Random
    & 0.500
    & 0.233 \\
    \rowcolor{myblue}
    Unmentioned
    & Omission
    & \textbf{\(\mathrm{PCS}\)}
    & \textbf{0.890}
    & \textbf{0.705} \\
    \bottomrule
    \end{tabular}
    }
\end{table}
\paragraph{Diagnosing evidence-verbalization mismatch.}
{We test whether pre-decoding VEX evidence (Sec.~\ref{subsec:vex}) identifies the two mismatch directions targeted by PatchGate.
Objects are split by baseline captions.
For mentioned objects, unsupported object mentions are ranked by \(1-\mathrm{PCS}\); for unmentioned objects, omitted visible objects are ranked by \(\mathrm{PCS}\).
Tab.~\ref{tab:ab_mismatch} shows that both rankings are well above random: AUROC reaches \(0.883/0.890\) versus \(0.500\), and AUPRC reaches \(0.707/0.705\) versus positive-rate baselines \(0.163/0.233\), respectively.
This indicates that VEX evidence diagnoses both hallucination- and omission-prone cases before decoding.}

\begin{table}[t]
    \centering
    \caption{{Performance and computational cost of VEX (Sec.~\ref{subsec:vex}) and external object inventories.}}
    \label{tab:ab_vex_effect}
    \adjustbox{max width=\linewidth}{
    \begin{tabular}{lcccc}
    \toprule
    \multirow{2}{*}{\textbf{Method}}
    & \multicolumn{2}{c}{\textbf{Metrics}}
    & \multicolumn{2}{c}{\textbf{Computational Cost}} \\
    \cmidrule(lr){2-3} \cmidrule(lr){4-5}
    & \textbf{Cover} $\uparrow$
    & \textbf{CHAIR} $\downarrow$
    & \textbf{Param.}
    & \textbf{Peak VRAM} \\
    \midrule
    LLaVA-v1.5-7B
    & 49.4 & 7.5
    & \textbf{7.0 B} & \textbf{14.6 GB} \\
    { + GroundingDINO}
    & 51.3 & 7.3
    & 7.2 B & 16.2 GB \\
    { + RAM++ (default)}
    & 50.9 & 7.2
    & 7.3 B & 16.9 GB \\
    { + RAM++ (top-\(10\))}
    & 50.9 & 7.5
    & 7.3 B & 16.9 GB \\
    \rowcolor{myblue}
    \textbf{ + VEX (Ours; Sec.~\ref{subsec:vex})}
    & \textbf{56.0} & \textbf{6.6}
    & \textbf{7.0 B} & 14.9 GB \\
    \bottomrule
    \end{tabular}
    }
\end{table}

\paragraph{Effect of VEX's internal object inventory.}
{Tab.~\ref{tab:ab_vex_effect} compares VEX with external models~\cite{liu2024groundingdinotiny, huang2025ram++} under the same decoding stage, VIED (Sec.~\ref{subsec:vied}).
Against all external-model variants, VEX improves Cover by at least \(9.2\%\), from the best external result \(51.3\) to \(56.0\), and reduces CHAIR by at least \(8.3\%\), from the best external result \(7.2\) to \(6.6\).
It achieves these gains while adding no additional parameters and only \(0.3\)GB peak VRAM over the LLaVA-v1.5-7B baseline.
This shows that object evidence extracted from the frozen VLM's internal representations provides a lightweight and more effective inventory for VIED's logit-space calibration than external object inventories.}

\begin{table}[t]
    \centering
    \caption{Effect of VIED components (Sec.~\ref{subsec:vied}) on AMBER.}
    \label{tab:ab_vied_effect}
    \adjustbox{max width=\linewidth}{
    \begin{tabular}{lcccccc}
    \toprule
    \multirow{2}{*}{\textbf{Method}} 
    & \multicolumn{2}{c}{\textbf{VIED}} 
    & \multicolumn{4}{c}{\textbf{Metrics}} \\
    \cmidrule(lr){2-3} \cmidrule(lr){4-7}
    & \textbf{ESI} 
    & \textbf{EDE} 
    & \textbf{Cover} $\uparrow$ 
    & \textbf{CHAIR} $\downarrow$ 
    & \textbf{Hal} $\downarrow$ 
    & \textbf{Cog} $\downarrow$ \\
    \midrule
    LLaVA-v1.5-7B & \xmark & \xmark & 49.4 & 7.5 & 31.4 & 3.6 \\
    { + PatchGate} & \cmark & \xmark & 54.8 & 6.7 & 34.6 & 3.9 \\
    { + PatchGate} & \xmark & \cmark & 50.3 & \textbf{6.2} & \textbf{29.5} & \textbf{2.8} \\
    \rowcolor{myblue} 
    \textbf{{ + PatchGate}} & \cmark & \cmark & \textbf{56.0} & 6.6 & 34.3 & 3.0 \\
    \bottomrule
    \end{tabular}
    }
\end{table}

\paragraph{Effect of VIED components.}
{ESI and EDE target different sides of the evidence-verbalization mismatch.
As shown in Tab.~\ref{tab:ab_vied_effect}, ESI alone mainly improves omission-related coverage, raising Cover from \(49.4\) to \(54.8\) (\(+10.9\%\)).
EDE alone most directly suppresses hallucination, reducing CHAIR from \(7.5\) to \(6.2\) (\(-17.3\%\)) and also improving Hal and Cog.
The full model is not best on hallucination-related diagnostics compared with EDE alone, but achieves the strongest Cover (\(56.0\)) while still reducing CHAIR from the baseline \(7.5\) to \(6.6\).
Thus, ESI and EDE are both needed to improve coverage and precision in a balanced way.}

\begin{table}[t]
    \centering
    \caption{Robustness of PatchGate across VLM backbones.}
    \label{tab:ab_robustness_vlm_backbone}
    \adjustbox{max width=\linewidth}{
    \begin{tabular}{lcccc}
    \toprule
    \textbf{Method}
    & \textbf{Cover} $\uparrow$
    & \textbf{CHAIR} $\downarrow$
    & \textbf{Hal} $\downarrow$
    & \textbf{Cog} $\downarrow$ \\
    \midrule
    LLaVA-v1.5-7B
    & 49.4 & 7.5 & \textbf{31.4} & 3.6 \\
    \rowcolor{myblue}
    \textbf{ + PatchGate}
    & \textbf{56.0} & \textbf{6.6} & 34.3 & \textbf{3.0} \\
    \midrule
    LLaVA-v1.5-13B
    & 50.8 & 6.5 & 30.8 & 3.1 \\
    \rowcolor{myblue}
    \textbf{ + PatchGate}
    & \textbf{52.8} & \textbf{5.7} & \textbf{25.7} & \textbf{2.2} \\
    \midrule
    Qwen2.5-VL-7B
    & 58.8 & 4.1 & 27.2 & 1.7 \\
    \rowcolor{myblue}
    \textbf{ + PatchGate}
    & \textbf{60.2} & \textbf{3.6} & \textbf{21.8} & \textbf{1.3} \\
    \midrule
    InstructBLIP-7B
    & 53.8 & 7.4 & 36.0 & 3.9\\
    \rowcolor{myblue}
    \textbf{ + PatchGate}
    & \textbf{54.5} & \textbf{6.9} & \textbf{35.4} & 3.9 \\
    \bottomrule
    \end{tabular}
    }
\end{table}
\paragraph{Robustness across VLM backbones.}
{PatchGate demonstrates consistent backbone-agnostic effectiveness across various VLM scales and families~\cite{liu2024llavaimproved, Qwen2.5-VL, dai2023instructblip}.
As shown in Tab.~\ref{tab:ab_robustness_vlm_backbone}, it improves Cover by ${+1.3}\%$ to ${+13.4}\%$ and reduces CHAIR by ${-6.8}\%$ to ${-12.3}\%$ across all evaluated backbones.
These results suggest that PatchGate does not rely on model-specific behaviors of a single backbone, but rather generalizes well across backbone scales and VLM families without fine-tuning or external detectors.}

\paragraph{Efficiency.}
{PatchGate is training-free and detector-free, but adds a small overhead from one VEX forward pass and lightweight VIED logit edits.
On LLaVA-v1.5-7B, runtime increases from \(2.92\)s to \(3.30\)s per image (\(1.13\times\)), and peak VRAM increases from \(14.6\)GB to \(14.9\)GB (\(1.02\times\)).
Thus, PatchGate mitigates both object hallucinations and omissions with modest overhead, while avoiding external visual modules, additional VLM queries, or model updates (see Appendix~\ref{app:ablation_study} for detailed efficiency results).}

\paragraph{Limitations.}
{PatchGate depends on the quality of the internal object inventory.
When VEX misses visible objects or assigns unreliable evidence, VIED has limited basis for correct inclusion or exclusion.
PatchGate is also limited to object-level logit calibration and does not directly address attribute, relation, counting, or higher-level scene hallucinations.
In some cases, this object-focused calibration may reduce descriptive attributes or modifiers compared with the baseline, affecting caption detail.
Further discussion is provided in Appendix~\ref{app:ablation_study}.}

\section{Conclusion} \label{sec:conclusion}
{We presented PatchGate, a training-free method for mitigating both object hallucinations and omissions in frozen Vision-Language Models.
Unlike prior training-free methods that mainly intervene on object words likely to emerge during generation, PatchGate first extracts prompt-free object evidence from internal patch representations and forms an intrinsic object inventory before decoding.
It then uses this inventory for two-sided logit calibration, promoting visually supported but under-verbalized objects and suppressing weakly supported object mentions.
Experiments on AMBER and POPE show that PatchGate improves object-level completeness and precision across both captioning and object-existence QA.
These results suggest that a frozen VLM's own patch-level evidence can serve as a practical grounding signal for more reliable object verbalization without external detectors or fine-tuning.}

\section*{Acknowledgements}
This work was partly supported by the KHIDI grant funded by the Korean government (MOHW) [No.RS-2025-02307233, No.RS-2026-25613012], the NRF or IITP grants funded by the Korean government (MSIT) [No.05-26-04-0094, No.RS-2026-25472075, No.RS-2025-02305581, No.RS-2025-25442338, and No.RS-2021-II211343], the ITIP grant funded by the Korean government (MOTIR) [No.RS-2026-25549946], the Research grant from SNU, and the Strategic Hub grant for International Research Collaboration of SNU.

\FloatBarrier
\appendix
\renewcommand{\thetable}{\Alph{table}}
\renewcommand{\thefigure}{\Alph{figure}}
\renewcommand{\thealgorithm}{\Alph{algorithm}}
\renewcommand{\theequation}{\Alph{equation}}
\setcounter{table}{0}
\setcounter{figure}{0}
\setcounter{algorithm}{0}
\setcounter{equation}{0}

% This work was partly supported by the KHIDI grant funded by the Korean government (MOHW) [No.RS-2025-02307233, No.RS-2026-25613012], the NRF or IITP grants funded by the Korean government (MSIT) [No.05-26-04-0094, No.RS-2026-25472075, No.RS-2025-02305581, No.RS-2025-25442338, and No.RS-2021-II211343], the ITIP grant funded by the Korean government (MOTIR) [No.RS-2026-25549946], the Research grant from SNU, and the Strategic Hub grant for International Research Collaboration of SNU. Kyungsu Kim is affiliated with the School of Transdisciplinary Innovations, Department of Biomedical Science, Interdisciplinary Program in Artificial Intelligence (IPAI), Medical Research Center, and AI Institute at SNU.

Kyungsu Kim is affiliated with the School of Transdisciplinary Innovations, Department of Biomedical Science, Interdisciplinary Program in Artificial Intelligence (IPAI), Medical Research Center, and AI Institute at SNU.

\appendix
\section*{Appendix}

\noindent\textbf{Overview.}
This supplementary material provides methodological details and additional analyses supporting the main paper.
Appendix~\ref{app:implementation_detail} details the methodology, algorithms, hyperparameters, and evaluation settings of PatchGate.
Additional quantitative and qualitative results, including extended baseline comparisons, are provided in Appendices~\ref{app:quantitative_results} and~\ref{app:qualitative_results}, respectively.
Appendix~\ref{app:ablation_study} examines key design choices, evidence aggregation, object-inventory sources, linguistic quality, and computational efficiency.
Finally, Appendix~\ref{app:limitations_future} discusses the limitations of the current object-level formulation and directions for future work.

\setcounter{table}{0}
\section{Implementation Details} \label{app:implementation_detail}

\subsection{Methodological Details} \label{app:methodological_details}

\subsubsection{VEX: Prompt-Free Object Inventory} \label{app:methodological_details_vex}

\paragraph{Image-only forward pass.}
PatchGate first applies VEX (Sec.~\ref{subsec:vex}) to read internal patch representations and extract an explicit image-conditioned object set before decoding.
VEX requires only one additional forward pass and neither uses a task-specific input prompt nor decodes any output tokens.
This differs from prior training-free methods~\cite{huang2024opera, leng2024vcd, zhao2025marine, jiang2025devils} that primarily derive intervention signals from object tokens likely to emerge during generation under a given prompt.
Because VEX is independent of both the downstream prompt and the model's emerging output tokens, it provides an image-specific signal that is available before generation begins.
The benchmark prompts provided by AMBER~\cite{wang2023amber} and POPE~\cite{li2023pope} are introduced only after the internal object inventory has been constructed, when VIED generates the final caption or binary answer.
We therefore refer to \(\mathcal{T}(I)\) as a prompt-free, image-conditioned object inventory extracted before generation.

\paragraph{Object vocabulary.}
We construct the object vocabulary \(\mathcal{V}_{\mathrm{obj}}\) from WordNet~\cite{miller1995wordnet} and use the same vocabulary in both VEX (Sec.~\ref{subsec:vex}) and VIED (Sec.~\ref{subsec:vied}).
The construction is described in four parts: (i) semantic class mapping from WordNet and part-of-speech tags, (ii) sense-frequency filtering after morphological normalization, (iii) broad vocabulary coverage followed by image-specific evidence selection, and (iv) tokenizer mapping that keeps only words represented by a single vocabulary token.

\textbf{(i) Semantic class mapping.}
Every WordNet synset carries a \emph{lexname}, one of its coarse \emph{supersenses} (\(26\) for nouns).
Tab.~\ref{tab:vobj_supersense} lists the mapping from lexname and POS to the five internal classes.
A word's noun, adjective, and verb senses are assigned to \textsc{object}, \textsc{scene}, \textsc{attribute}, \textsc{action}, or \textsc{body} according to this mapping.

\begin{table}[htbp]
\centering
\caption{WordNet supersense (lexname) and part-of-speech (POS) mapping used to construct \(\mathcal{V}_{\mathrm{obj}}\). Only \textsc{object} and \textsc{scene} classes are included.}
\label{tab:vobj_supersense}
\adjustbox{max width=\linewidth}{
\begin{tabular}{llcc}
\toprule
\textbf{POS} & \textbf{WordNet lexname(s)} & \textbf{Class} & \textbf{Included} \\
\midrule
noun & \shortstack[l]{\texttt{noun.artifact}, \texttt{noun.animal}, \texttt{noun.food},\\ \texttt{noun.person}, \texttt{noun.plant}} & \textsc{object} & \cmark \\
noun & \texttt{noun.location}, \texttt{noun.object}, \texttt{noun.substance} & \textsc{scene} & \cmark \\
noun & \texttt{noun.act}, \texttt{noun.event} & \textsc{action} & \xmark \\
noun & \texttt{noun.body} & \textsc{body} & \xmark \\
noun & all other \texttt{noun.*} & --- & \xmark \\
adj. (\texttt{a}/\texttt{s}) & all adjective synsets & \textsc{attribute} & \xmark \\
verb (\texttt{v}) & all verb synsets & \textsc{action} & \xmark \\
\bottomrule
\end{tabular}}
\end{table}

\textbf{(ii) Sense-frequency filtering.}
The construction uses standard WordNet resources together with SemCor sense frequencies.
Each word \(w\) is lowercased and lemmatized to \(n=\mathrm{lemma}(w,\textsc{n})\); we then accumulate the SemCor occurrence count of each sense associated with the normalized form into the per-class tally \(\mathrm{cnt}_k(w)\) over the five classes \(\mathcal{K}=\{\textsc{object},\textsc{scene},\textsc{attribute},\textsc{action},\textsc{body}\}\).
The word enters \(\mathcal{V}_{\mathrm{obj}}\) only when its frequency-dominant class is \textsc{object} or \textsc{scene} (see Tab.~\ref{tab:vobj_supersense}).

\textbf{(iii) Broad vocabulary coverage.}
\(\mathcal{V}_{\mathrm{obj}}\) is deliberately not required to be a precise object lexicon: it only has to be a \emph{superset} of the nameable object and scene words that could be read off a patch. The downstream VEX stages perform the actual image-specific selection. Thus, a word that is over-included in \(\mathcal{V}_{\mathrm{obj}}\) but carries little visual evidence receives a low \(\mathrm{PCS}\) and is unlikely to enter \(\mathcal{T}(I)\).
If its decoding logit later becomes high, the low \(\mathrm{PCS}\) also produces a stronger EDE penalty during VIED.

\textbf{(iv) Tokenizer mapping.} 
Finally, we intersect the WordNet-derived object and scene lexicon with the VLM backbone's own subword vocabulary.
We iterate over every token ID in the tokenizer, decode it to its surface string, and retain the token if the string is alphabetic, contains at least three characters, and passes \(\textsc{PrimaryCat}\).
Because the enumeration visits one ID at a time, every retained entry is by construction a single token, so its probability is exactly the logit-lens softmax coordinate at that ID.
The output is a fixed ID-to-word table defining \(\mathcal{V}_{\mathrm{obj}}\).
For LLaVA-1.5-7B's \(32\)k-token vocabulary~\cite{liu2024llavaimproved}, this procedure yields \(1{,}560\) object and scene tokens, which are reused unchanged across all images.

\paragraph{Object evidence scoring.}
Although the image-specific candidate set \(\mathcal{C}(I)\) contains only top-1 vocabulary readouts belonging to \(\mathcal{V}_{\mathrm{obj}}\), VEX computes the patch-confidence score \(\mathrm{PCS}(c)\) for every object \(c \in \mathcal{V}_{\mathrm{obj}}\) as the maximum full-vocabulary softmax probability over all patch-layer pairs in the selected late-layer range.
Thus, \(\mathcal{C}(I)\) determines which objects are eligible to enter the final inventory \(\mathcal{T}(I)\), while the continuous PCS values are retained for evidence-dependent exclusion over the entire object vocabulary during VIED (Sec.~\ref{subsec:vied}).
We use maximum aggregation to preserve strong localized evidence that may appear within a small image region, since mean aggregation can dilute such evidence when most patches are unrelated to a given object, while relying on a single fixed layer can be brittle when object information becomes lexically readable at different late layers.
We analyze these aggregation choices in Sec.~\ref{app:ablation_study}.

\begin{figure}[t]
    \centering
    \includegraphics[width=\linewidth]{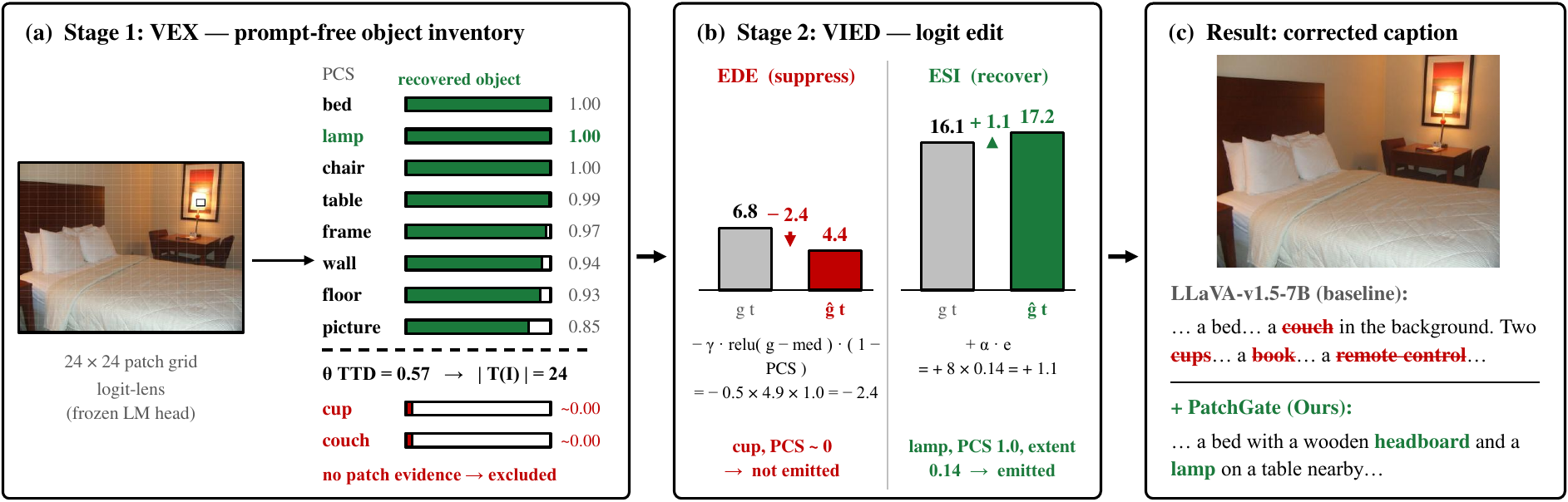}
    \caption{\textbf{End-to-end PatchGate inference for open-ended captioning.}
    VEX (Sec.~\ref{subsec:vex}) selects the largest-gap cutoff at \(\theta_{\mathrm{TTD}}=0.57\), producing an internal object inventory that retains supported objects (\emph{bed}, \emph{lamp}, \emph{chair}, and \emph{table}) while excluding unsupported objects (\emph{cup} and \emph{couch}). During VIED (Sec.~\ref{subsec:vied}), EDE reduces the \emph{cup} logit from \(6.8\) to \(4.3\), whereas ESI increases the omitted \emph{lamp} logit from \(16.1\) to \(17.2\), causing it to be generated.}
    \label{fig:worked_example_captioning}
\end{figure}

\paragraph{Adaptive object selection.}
After computing the PCS values, VEX constructs the final object inventory \(\mathcal{T}(I)\) using an image-adaptive threshold motivated by the largest-score-gap selection strategy of TTD~\cite{jo2024ttd} (see Fig.~\ref{fig:worked_example_captioning}).
Candidates in \(\mathcal{C}(I)\) with scores below the evidence floor \(\tau_{\mathrm{floor}}\) are first removed because their visual support is too weak to justify inclusion in \(\mathcal{T}(I)\), preventing low-confidence objects from entering the inventory solely due to relative gaps among low PCS values.
Let \(c_1,\ldots,c_m\) denote the remaining candidates ordered by descending PCS, with corresponding scores \(s_1 \geq s_2 \geq \cdots \geq s_m\).
We measure the decrease between each pair of adjacent scores as
\begin{equation}
\delta_i = s_i - s_{i+1},
\qquad i=1,\ldots,m-1.
\label{eq:adaptive_gap}
\end{equation}
A large \(\delta_i\) indicates a separation between the higher-scoring candidates above position \(i\) and the lower-scoring candidates below it.
For \(m \geq 3\), we exclude the final gap \(\delta_{m-1}\), which compares the two lowest-scoring candidates, because a single unusually weak candidate can make this gap large without representing the main separation in the score distribution.
Among the remaining gaps, we choose the largest one and define the adaptive threshold as
\begin{equation}
k^\star
=
\operatorname*{arg\,max}_{1\le i\le m-2}
\delta_i,
\qquad
\theta_{\mathrm{TTD}}(I)
=
\frac{s_{k^\star}+s_{k^\star+1}}{2}.
\label{eq:adaptive_threshold}
\end{equation}
When multiple positions have the same gap, the smallest index is selected.
The final object inventory is then defined as
\begin{equation}
\mathcal{T}(I)
=
\left\{
c \in \mathcal{C}(I)
\;\middle|\;
\mathrm{PCS}(c) \geq \tau_{\mathrm{floor}},
\;
\mathrm{PCS}(c) > \theta_{\mathrm{TTD}}(I)
\right\}.
\label{eq:adaptive_inventory}
\end{equation}
If \(m=0\), the inventory is empty.
If \(m=1\), the sole candidate is included, while for \(m=2\), the single available gap is used without tail-gap exclusion.

\subsubsection{VIED: Task-Specific Decoding}
\label{app:methodological_details_vied}
\paragraph{Task-specific inventory usage.}
Once the object inventory \(\mathcal{T}(I)\) has been constructed, PatchGate adapts how it is used according to the output space. For open-ended captioning, ESI and EDE directly adjust object-token logits (Eqs.~\ref{eq:esi_bonus} and~\ref{eq:ede_penalty}). Binary QA, by contrast, operates in the answer-token space, using inventory-conditioned classifier-free guidance to adjust the relative logits of ``Yes'' and ``No''.
This task-specific formulation allows the same object inventory to support both free-form generation and binary decisions.

\paragraph{Open-ended captioning.}
The PCS and extent values produced by VEX are precomputed once and reused throughout decoding.
At each generation step \(t\), ESI and EDE are computed with respect to the same unmodified base-logit vector \(g_t\), and their corrections are added simultaneously before token selection.
After each generated token, VIED updates the object-mention set \(\mathcal{M}_{<t}\) according to the object-realization rule.
Once an object in inventory is marked as realized, its ESI bonus (Eq.~\ref{eq:esi_bonus}) is disabled for subsequent steps, while EDE penalty (Eq.~\ref{eq:ede_penalty}) remains active and the frozen decoder may still generate the object again through its original logits.

\paragraph{Binary QA.}
\ej{Because binary QA expresses its final decision through the ``Yes'' and ``No'' tokens, which are not directly affected by the object-token updates of ESI and EDE, we adapt the image-grounded classifier-free guidance formulation of MARINE~\cite{zhao2025marine}.
Given the original image and benchmark query, the unconditioned inference produces the logit vector \(g_t^{\mathrm{uncond}}\), whereas the conditioned inference additionally incorporates the object inventory \(\mathcal{T}(I)\) as image-grounded context and produces \(g_t^{\mathrm{cond}}\).
At the answer step \(t\), these two signals are combined to obtain the guided logit vector
\begin{equation}
g_t^{\mathrm{QA}}
=
\lambda_{\mathrm{QA}}g_t^{\mathrm{cond}}
+
\left(1-\lambda_{\mathrm{QA}}\right)g_t^{\mathrm{uncond}}.
\label{eq:qa_guidance}
\end{equation}
Here, the QA guidance strength \(\lambda_{\mathrm{QA}}\in[0,1]\) controls the contribution of the inventory-conditioned logits relative to the unconditioned logits, and the final answer is selected by comparing the ``Yes'' and ``No'' entries of \(g_t^{\mathrm{QA}}\).
This enables bidirectional correction: the conditioned signal can raise the relative ``Yes'' logit when the unconditioned inference incorrectly answers ``No'' for a visually supported object, while shifting the decision toward ``No'' when the unconditioned inference produces a hallucinated affirmative answer for an unsupported object.}

\begin{algorithm}[t]
\caption{VEX: prompt-free object inventory $\mathcal{T}(I)$}
\label{alg:vex}
\begin{algorithmic}[1]
\REQUIRE image $I$ (no text prompt), frozen VLM with unembedding $W_U$ and $\mathrm{RMSNorm}$, later-layer set $\mathcal{L}$, object vocabulary $\mathcal{V}_{\mathrm{obj}}$, floor $\tau_{\mathrm{floor}}$
\STATE $\{h^{(\ell)}_p\} \leftarrow \text{VLM.\textsc{forward}}(I)$ \hfill$\triangleright$ image only, single forward pass, no decoding
\STATE \textbf{// Step 1: patch-to-vocabulary readout}
\FOR{each patch $p\in\{1,\dots,N\}$, layer $\ell\in\mathcal{L}$}
    \STATE $q^{(\ell)}_p(v) \leftarrow \mathrm{softmax}\!\big(W_U\,\mathrm{RMSNorm}(h^{(\ell)}_p)\big)[v]$ \hfill$\triangleright$ logit lens
\ENDFOR
\STATE $\mathcal{C}(I) \leftarrow \{\arg\max_{v} q^{(\ell)}_p(v) : \ell\in\mathcal{L},\, p\}\cap\mathcal{V}_{\mathrm{obj}}$
\STATE \textbf{// Step 2: object evidence scoring}
\STATE $\mathrm{PCS}(c)\leftarrow\max_{p}\max_{\ell\in\mathcal{L}} q^{(\ell)}_p(c)$ \quad for each {$c\in\mathcal{V}_{\mathrm{obj}}$}
\STATE \textbf{// Step 3: adaptive object selection}
\STATE $\mathcal{T}(I) \leftarrow \{c\in\mathcal{C}(I): \mathrm{PCS}(c)\ge\tau_{\mathrm{floor}}, \ \mathrm{PCS}(c)>\theta_{\mathrm{TTD}}(I)\}$ \hfill$\triangleright$ largest-gap cut, Eq.~\ref{eq:adaptive_threshold}
\STATE extent $e(c)\in[0,1]$ for $c\in\mathcal{T}(I)$ by Eq.~\ref{eq:esi_bonus}
\STATE \textbf{return} inventory $\mathcal{T}(I)$, evidence $\mathrm{PCS}(\cdot)$, extent $e(\cdot)$
\end{algorithmic}
\end{algorithm}

\subsection{PatchGate Algorithm} \label{app:patchgate_algorithm}
For reproducibility, Algorithms~\ref{alg:vex} and~\ref{alg:vied} summarize the complete inference procedure of PatchGate, while the full implementation code and configuration files are provided in a separately submitted code archive.

\paragraph{{VEX.}} 
Stage 1 of PatchGate, VEX (Sec.~\ref{subsec:vex}), extracts the prompt-free object inventory \(\mathcal{T}(I)\) and computes the PCS and extent scores through a single image-only forward pass without any task-specific prompt, as summarized in Algorithm~\ref{alg:vex}.

\paragraph{VIED.}
Stage 2 of PatchGate, VIED (Sec.~\ref{subsec:vied}), uses the extracted inventory, PCS, and extent scores during greedy decoding to suppress unsupported object mentions through EDE and recover omitted visible objects through ESI, as summarized in Algorithm~\ref{alg:vied}.

\begin{algorithm}[t]
\caption{VIED: inclusion/exclusion decoding}
\label{alg:vied}
\begin{algorithmic}[1]
\REQUIRE image $I$, prompt $x$, inventory $\mathcal{T}(I)$, evidence $\mathrm{PCS}(\cdot)$, extent $e(\cdot)$, object vocabulary $\mathcal{V}_{\mathrm{obj}}$, strengths $\alpha_{\mathrm{esi}}$, $\gamma_{\mathrm{ede}}$, maximum output length $T_{\max}$
\STATE $y \leftarrow [\,]$; \; $\mathcal{M} \leftarrow \varnothing$ \hfill$\triangleright$ output tokens; realized inventory objects
\REPEAT
    \STATE $g_t \leftarrow \text{VLM}(x, I, y)$; \quad $\mathrm{med}_t \leftarrow \mathrm{median}_{v\in\mathcal{V}}\, g_t[v]$ \hfill$\triangleright$ base logits, neutral level
    \STATE $\hat{g}_t \leftarrow g_t$
    \FOR{each object $c \in \mathcal{V}_{\mathrm{obj}}$}
        \STATE $\hat{g}_t[c] \leftarrow \hat{g}_t[c] - \gamma_{\mathrm{ede}}\,\mathrm{relu}\!\big(g_t[c]-\mathrm{med}_t\big)\big(1-\mathrm{PCS}(c)\big)$ \hfill$\triangleright$ EDE
    \ENDFOR
    \FOR{each $c \in \mathcal{T}(I)$ with $c\notin\mathcal{M}$}
        \STATE $\hat{g}_t[c] \leftarrow \hat{g}_t[c] + \alpha_{\mathrm{esi}}\, e(c)$ \hfill$\triangleright$ ESI (fix-once)
    \ENDFOR
    \STATE $v_t \leftarrow \arg\max_{v\in\mathcal{V}}\hat{g}_t[v]$; \;
    $y.\textsc{append}(v_t)$
    \STATE \textbf{if} $v_t$ realizes an object $c\in\mathcal{T}(I)$
    \textbf{then} $\mathcal{M}\leftarrow\mathcal{M}\cup\{c\}$
\UNTIL{$v_t=\texttt{EOS}$ or $|y|=T_{\max}$}
\STATE \textbf{return} generated caption $y$
\end{algorithmic}
\end{algorithm}

\begin{figure}[t]
    \centering
    \includegraphics[width=1.0\linewidth]{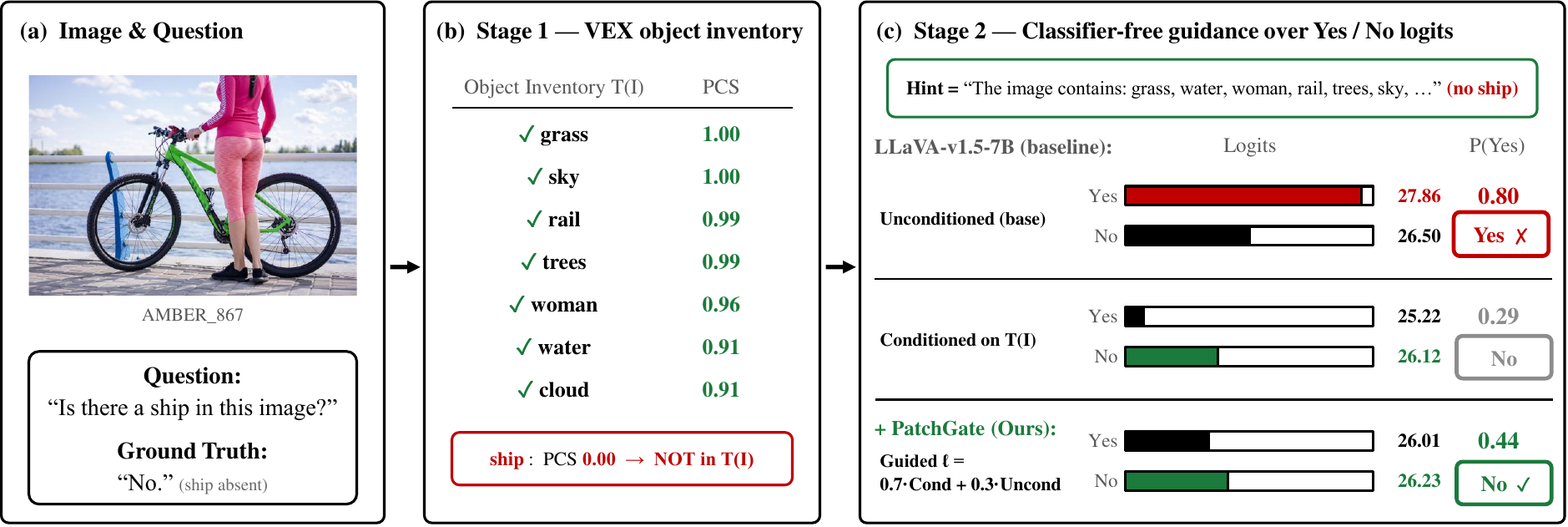}
    \caption{\textbf{End-to-end PatchGate inference for binary QA.}
    PatchGate constructs the object inventory \(\mathcal{T}(I)\), in which \emph{ship} is absent due to its negligible visual support (\(\mathrm{PCS}\approx0.00\)). While the unconditioned LLaVA-v1.5-7B logits yield a hallucinated ``Yes'' answer, inventory-conditioned guidance raises the ``No'' logit above ``Yes'', yielding the correct answer ``No''.}
    \label{fig:worked_example_binaryqa}
\end{figure}

\paragraph{Worked example: open-ended captioning (Fig.~\ref{fig:worked_example_captioning}).} \label{app:worked_captioning}
To provide a more concrete understanding of PatchGate, we trace its end-to-end inference procedure on AMBER image 209~\cite{wang2023amber}, which depicts a hotel room whose base caption generated by LLaVA-1.5-7B~\cite{liu2024llavaimproved} hallucinates a \underline{couch}, two \underline{cups}, a \underline{book}, and a \underline{remote control}, while omitting the visible \underline{lamp} and \underline{headboard} (see Fig.~\ref{fig:worked_example_captioning}).
All reported values are the actual PCS, extent scores, and decoding logits, with \(\mathcal{L}=\{22,\dots,32\}\), \(\alpha_{\mathrm{esi}}=8\), and \(\gamma_{\mathrm{ede}}=0.5\).

\textbf{(i) VEX (Alg.~\ref{alg:vex}).} 
PatchGate first projects every visual patch into the vocabulary space through the logit lens~\cite{geva2022transformer} (VEX Step~1; Sec.~\ref{subsec:vex}) and computes the patch-confidence score \(\mathrm{PCS}(c)\) for every object \(c\in\mathcal{V}_{\mathrm{obj}}\) from the strongest response across the selected patches and layers (VEX Step~2).
The room's visually supported objects form a high-evidence cluster, including
\(\underline{\text{bed}}\,(1.00)\),
\(\underline{\text{lamp}}\,(1.00)\),
\(\underline{\text{chair}}\,(1.00)\), and
\(\underline{\text{table}}\,(0.99)\),
whereas the hallucinated \underline{cup} has \(\mathrm{PCS}(\text{cup})=0.00\), and the \underline{couch} receives approximately zero patch evidence.
PatchGate then selects the largest eligible interior gap at \(\theta_{\mathrm{TTD}}=0.57\) (VEX Step~3), yielding an inventory of size \(|\mathcal{T}(I)|=24\): the \underline{lamp} is retained with extent \(e(\text{lamp})=0.14\), whereas the \underline{cup} and \underline{couch} fall below the threshold and are excluded.

\textbf{(ii) VIED (Alg.~\ref{alg:vied}).} 
When the language model assigns a high logit to the visually unsupported \underline{cup}, its base logit reaches \(g_t[\text{cup}]=6.8\), above the neutral median \(\mathrm{med}_t=1.9\).
Because \(\mathrm{PCS}(\text{cup})\approx0\), PatchGate applies EDE (Sec.~\ref{subsec:vied}) and subtracts
\(\gamma_{\mathrm{ede}}\mathrm{relu}(g_t[\text{cup}]-\mathrm{med}_t)(1-\mathrm{PCS}(\text{cup}))=0.5\cdot4.9\cdot1.0\approx\rev{2.5}\),
thereby reducing the \underline{cup} logit to
\(\hat{g}_t[\text{cup}]=6.8-\rev{2.5}\approx\rev{4.3}\), which prevents it from being generated, while the same mechanism suppresses the other unsupported objects.
For the visually supported but omitted \underline{lamp}, PatchGate applies ESI (Sec.~\ref{subsec:vied}) and adds
\(\alpha_{\mathrm{esi}}e(\text{lamp})=8\cdot0.14=1.12\approx1.1\)
at every decoding step until it is generated, increasing its logit to
\(\hat{g}_t[\text{lamp}]=16.1+1.12=17.22\approx17.2\), so the \underline{lamp} becomes the argmax token.
The resulting caption excludes the hallucinated objects and mentions the visible \underline{lamp} and \underline{headboard}.

\paragraph{Worked example: binary QA (Fig.~\ref{fig:worked_example_binaryqa}).} \label{app:worked_binaryqa}
\ej{On an AMBER example asking whether a \underline{ship} is present, the queried object receives negligible visual support (\(\mathrm{PCS}\approx0.00\)) and is therefore absent from the resulting inventory \(\mathcal{T}(I)\), which instead contains visually supported objects such as \underline{sky}, \underline{trees}, and \underline{water}.
Nevertheless, the unconditioned LLaVA-v1.5-7B logits favor ``Yes'' over ``No'' (\(27.86\) vs.\ \(26.50\)), yielding a hallucinated affirmative answer.
Conditioning on \(\mathcal{T}(I)\) reverses this preference, producing ``Yes'' and ``No'' logits of \(25.22\) and \(26.12\), respectively.
With \(\lambda_{\mathrm{QA}}=0.7\), these conditioned logits are combined with the unconditioned logits as
\(\text{``Yes'' logit}=0.7\times25.22+0.3\times27.86\approx26.01\) and
\(\text{``No'' logit}=0.7\times26.12+0.3\times26.50\approx26.23\).
Consequently, the ``No'' logit becomes higher than the ``Yes'' logit, allowing PatchGate to produce the correct answer ``No''.}

\subsection{Experimental Details} \label{app:experimental_details}

Tab.~\ref{tab:ap_main_settings} summarizes the model, main hyperparameters, decoding settings, and computational environment used in our experiments.
Additional details on baseline implementations, prompt protocols, binary QA decoding, and benchmark-specific evaluation are provided below.

\begin{table}[ht]
    \centering
    \caption{Main hyperparameters and implementation settings used for PatchGate.}
    \label{tab:ap_main_settings}
    \adjustbox{max width=\linewidth}{
    \begin{tabular}{ll}
    \toprule
    \textbf{Setting} & \textbf{Value} \\
    \midrule
    \rowcolor{gray!15}
    \multicolumn{2}{l}{\textbf{PatchGate setup}} \\
    \midrule
    Backbone & LLaVA-v1.5-7B \\
    HuggingFace model ID & \texttt{llava-hf/llava-1.5-7b-hf} \\
    VEX layer range \(\mathcal{L}\) & \(\{22,\ldots,32\}\) \\
    Evidence floor \(\tau_{\mathrm{floor}}\) & \(0.02\) \\
    ESI strength \(\alpha_{\mathrm{esi}}\) & \(8\) \\
    EDE strength \(\gamma_{\mathrm{ede}}\) & \(0.5\) \\
    QA guidance strength \(\lambda_{\mathrm{QA}}\) & \(0.7\) \\
    Decoding & Greedy \\
    \texttt{do\_sample} & \texttt{False} \\
    Random seed & Not used (deterministic) \\
    Temperature & Not used \\
    \midrule
    \rowcolor{gray!15}
    \multicolumn{2}{l}{\textbf{Sampling-based baselines}} \\
    \midrule
    Random seed & \(42\) \\
    Temperature & \(1.0\) \\
    \midrule
    \rowcolor{gray!15}
    \multicolumn{2}{l}{\textbf{Hardware and software}} \\
    \midrule
    GPU & 1 NVIDIA RTX A6000 \rev{48GB} \\
    Python & 3.10.12 \\
    PyTorch & 2.1.2 \\
    Transformers & 4.45.2 \\
    \bottomrule
    \end{tabular}
    }
\end{table}

\paragraph{Baselines.}
We compare PatchGate with the frozen LLaVA-v1.5-7B backbone~\cite{liu2024llavaimproved} and representative training-free methods: OPERA~\cite{huang2024opera}, VCD~\cite{leng2024vcd}, Devils-in-Mid.~\cite{jiang2025devils}, MARINE~\cite{zhao2025marine}, ProjectAway~\cite{jiang2025projectaway}, ILVAD~\cite{Xie2026ilvad}, SHIELD~\cite{huang2026shield}, and PND~\cite{jiang2026pnd}.
We use the official implementations whenever available; because PND had no public implementation at the time of our experiments, we reimplemented it following the paper.
All methods are evaluated under the same benchmark protocols while preserving their method-specific inference procedures.

\paragraph{Benchmarks.}
We evaluate PatchGate on two object-hallucination benchmarks, AMBER~\cite{wang2023amber} and POPE~\cite{li2023pope}.
AMBER provides a generative task with \(1{,}004\) free-form image-description samples and a discriminative binary QA task covering object existence, attribute, and relation hallucinations.
Its object-existence subset asks only about plausible but absent objects, and therefore all ground-truth answers are ``No.''
This subset evaluates absent-object rejection but does not measure the complementary ability to affirm visible objects.
In contrast, POPE contains both present-object questions with ``Yes'' answers and absent-object questions with ``No'' answers under random, popular, and adversarial negative-object sampling.
It therefore evaluates both visible-object acceptance and absent-object rejection.
For both benchmarks, we follow the official evaluation protocols and use the provided annotations and evaluation scripts.

\begin{figure}[t]
    \centering
    \includegraphics[width=1.0\linewidth]{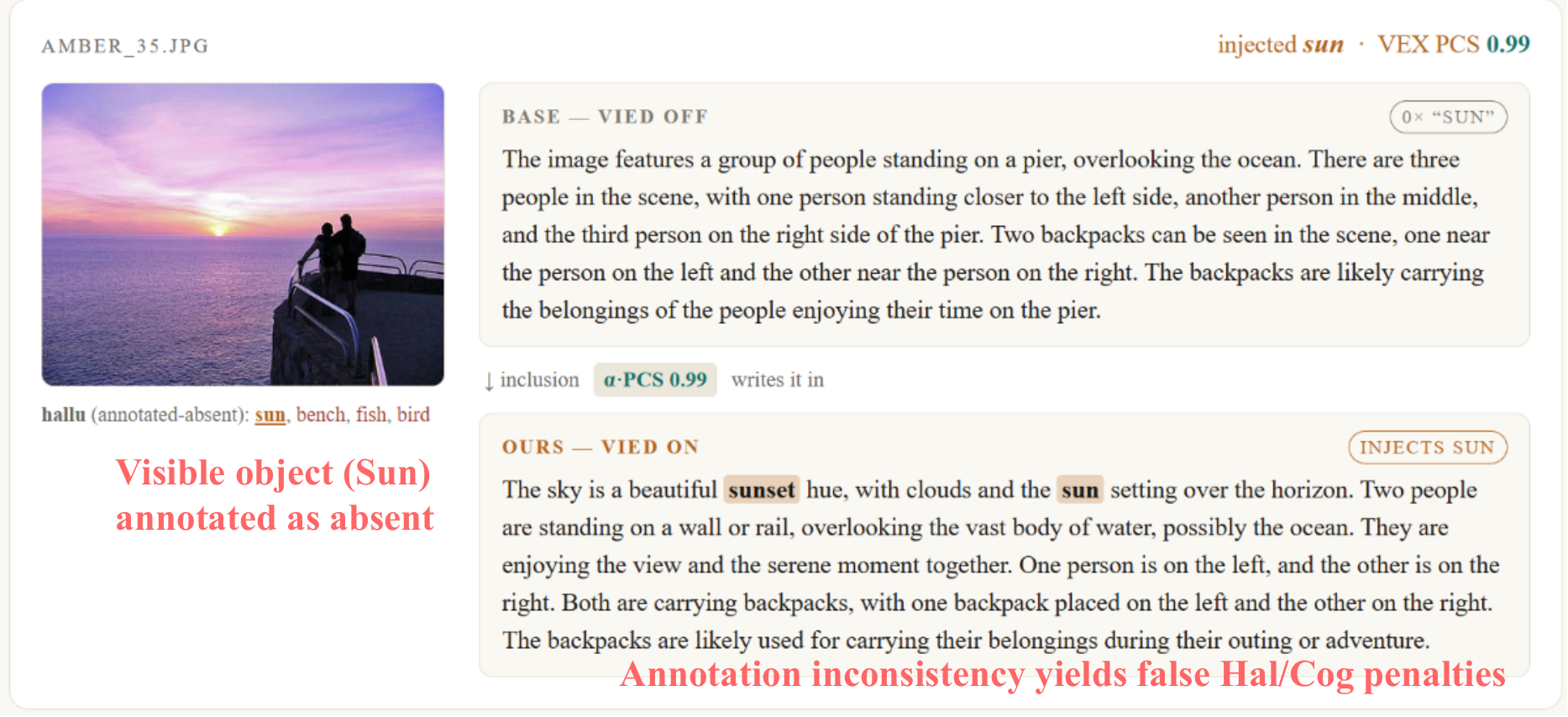}
    \caption{\textbf{Limitations of Hal and Cog.}
    Although the sun is visibly present, AMBER annotates it as absent and includes it in the image-specific hallucinatory target set. PatchGate correctly recovers \emph{sun}, but this mention is penalized by both Hal and Cog.}
    \label{fig:hal_cog_fail}
\end{figure}

\paragraph{Metrics.} We evaluate generative captioning and binary QA using the official metrics provided by AMBER~\cite{wang2023amber} and POPE~\cite{li2023pope}.

\textbf{Generative captioning.}
For a generated response \(R\), the official AMBER evaluator first extracts its nouns to obtain \(R_{\mathrm{obj}}\), and then retains only those included in the complete AMBER object list \(X_{\mathrm{obj}}\), yielding \(R'_{\mathrm{obj}}=R_{\mathrm{obj}}\cap X_{\mathrm{obj}}\).
Let \(A_{\mathrm{obj}}\) denote the visible objects annotated for the image and \(H_{\mathrm{obj}}\) denote its image-specific hallucinatory target objects.
Following the official AMBER definitions~\cite{wang2023amber}, the four generative metrics are
\begin{equation}
\begin{aligned}
\mathrm{Cover}(R)
&=
\frac{
\left|R'_{\mathrm{obj}}\cap A_{\mathrm{obj}}\right|
}{
\left|A_{\mathrm{obj}}\right|
},
\\[2mm]
\mathrm{CHAIR}(R)
&=
1-
\frac{
\left|R'_{\mathrm{obj}}\cap A_{\mathrm{obj}}\right|
}{
\left|R'_{\mathrm{obj}}\right|
},
\\[2mm]
\rev{\mathrm{Hal}(R)}
&=
\rev{\mathbf{1}\!\left[\mathrm{CHAIR}(R)>0\right]},
\\[2mm]
\mathrm{Cog}(R)
&=
\frac{
\left|R'_{\mathrm{obj}}\cap H_{\mathrm{obj}}\right|
}{
\left|R'_{\mathrm{obj}}\right|
}.
\end{aligned}
\label{eq:amber_generative_metrics}
\end{equation}
Cover measures the proportion of annotated visible objects mentioned in the response, while CHAIR measures the proportion of generated object mentions that are unsupported by the image.
Hal indicates whether a response contains at least one unsupported object mention, whereas Cog measures the proportion of AMBER's predefined, image-specific hallucinatory targets generated in the response.
The four metrics are reported over all \(1{,}004\) generative samples.
Higher Cover is better, whereas lower CHAIR, Hal, and Cog are better.

\textbf{Limitations of Hal and Cog.}
Although Hal and Cog provide complementary hallucination diagnostics, they are sensitive to the coverage and consistency of AMBER's object annotations.
Hal is a caption-level binary metric that marks an entire response as hallucinated when even one extracted object mention is not matched to the annotated visible-object set.
Thus, a single annotation mismatch can change Hal from \(0\) to \(1\), while affecting the mention-level CHAIR score more gradually.
Cog additionally depends on AMBER's predefined, image-specific hallucinatory target set.
As shown in Fig.~\ref{fig:hal_cog_fail}, the clearly visible \emph{sun} is absent from the visible-object annotations and instead included in the hallucinatory target set.
Consequently, PatchGate's visually grounded recovery of \emph{sun} is penalized by both Hal and Cog.
Such an annotation mismatch is not directly penalized by Cover, but the metric cannot credit the recovered object; CHAIR instead counts the unmatched mention as unsupported, although its effect remains proportional rather than caption-level.
We therefore use Cover and CHAIR as the primary complementary measures of object coverage and mention-level hallucination, while interpreting Hal and Cog as additional diagnostics that can be more sensitive to annotation inconsistencies.

\textbf{Discriminative QA.}
For AMBER discriminative QA and POPE, we report Accuracy (Acc.), Precision (P.), Recall (R.), and F1, computed as follows:
\begin{equation}
\begin{aligned}
\mathrm{Acc.}
&=
\frac{
\mathrm{TP}+\mathrm{TN}
}{
\mathrm{TP}+\mathrm{TN}+\mathrm{FP}+\mathrm{FN}
},
\\
\mathrm{P.}
&=
\frac{
\mathrm{TP}
}{
\mathrm{TP}+\mathrm{FP}
},
\\
\mathrm{R.}
&=
\frac{
\mathrm{TP}
}{
\mathrm{TP}+\mathrm{FN}
},
\\
\mathrm{F1}
&=
\frac{
2\,\mathrm{P}\times\mathrm{R}
}{
\mathrm{P}+\mathrm{R}
}.
\end{aligned}
\label{eq:discriminative_metrics}
\end{equation}
The two benchmarks follow different positive-label conventions.
Following the official AMBER evaluator, ``No'' is treated as the positive label.
Its object-existence subset contains only questions about absent objects with the ground-truth answer ``No'', and therefore isolates absent-object rejection.
The full discriminative results in Tab.~\ref{tab:ap_amber}, however, also include attribute and relation questions.
Under this convention, Precision measures the proportion of predicted ``No'' answers that are correct, while Recall measures the proportion of ground-truth ``No'' cases correctly answered ``No''.

In contrast, the official POPE evaluator~\cite{li2023pope} treats ``Yes'' as the positive label.
A false positive is an absent-object question incorrectly answered ``Yes'', corresponding to hallucination, while a false negative is a visible-object question incorrectly answered ``No'', corresponding to omission.
Precision therefore reflects resistance to hallucinated positive answers, whereas Recall reflects omission recovery.
Higher values indicate better performance for all four metrics.

\begin{table}[p]
    \centering
    \caption{Full AMBER results~\cite{wang2023amber} across different VLM backbones. We include all training-free baselines evaluated on each backbone.
    For Cover, CHAIR, P., and R., colored values show absolute changes from the corresponding backbone baseline (green: improvement, red: degradation). The best results within each backbone are in \textbf{bold}.}
    \label{tab:ap_amber}
    \resizebox{1\linewidth}{!}{
    \begin{tabular}{lcccccccc}
    \toprule
    \multirow{2}{*}{\textbf{Method}}
    & \multicolumn{4}{c}{\textbf{Generative Task}}
    & \multicolumn{4}{c}{\textbf{Discriminative Task}} \\
    \cmidrule(lr){2-5} \cmidrule(lr){6-9}
    & \textbf{Cover} $\uparrow$
    & \textbf{CHAIR} $\downarrow$
    & \textbf{Hal} $\downarrow$
    & \textbf{Cog} $\downarrow$
    & \textbf{Acc.} $\uparrow$
    & \textbf{P.} $\uparrow$
    & \textbf{R.} $\uparrow$
    & \textbf{F1} $\uparrow$ \\
    \midrule
    \rowcolor{gray!15}
    \multicolumn{9}{l}{\textbf{[CVPR'24] LLaVA-v1.5-7B}} \\
    \midrule
    {[CVPR'24] LLaVA-v1.5-7B}
    & 49.4 & 7.5 & 31.4 & 3.6
    & 72.0 & 92.5 & 62.9 & 74.9 \\
    { + [CVPR'24] OPERA}
    & \vbad{48.8}{-0.6}
    & \vimp{6.6}{-0.9}
    & 28.3 & 2.9
    & \textbf{76.4}
    & \vimp{92.7}{+0.2}
    & \vimp{69.9}{+7.0}
    & \textbf{79.7} \\
    { + [CVPR'24] VCD}
    & \vimp{51.7}{+2.3}
    & \vbad{8.9}{+1.4}
    & 39.7 & 4.4
    & 71.1
    & \vbad{91.4}{-1.1}
    & \vbad{62.2}{-0.7}
    & 74.0 \\
    { + [CVPR'25] Devils-in-Mid.}
    & \vimp{50.3}{+0.9}
    & \bvimp{3.3}{-4.2}
    & 20.0 & \textbf{1.2}
    & 50.5
    & \vimp{94.5}{+2.0}
    & \vbad{26.8}{-36.1}
    & 41.8 \\
    { + [ICML'25] MARINE}
    & \vimp{49.9}{+0.5}
    & \vimp{6.0}{-1.5}
    & 27.3 & 2.9
    & 72.4
    & \bvimp{94.8}{+2.3}
    & \vbad{61.7}{-1.2}
    & 74.7 \\
    { + [ICLR'25] ProjectAway}
    & \vneu{49.4}{+0.0}
    & \vbad{9.2}{+1.7}
    & 37.6 & 5.0
    & 70.6
    & \vneu{92.5}{+0.0}
    & \vbad{60.6}{-2.3}
    & 73.2 \\
    { + [ICML'26] ILVAD}
    & \vbad{44.9}{-4.5}
    & \vimp{4.1}{-3.4}
    & \textbf{19.0} & 1.3
    & 75.4
    & \vbad{90.5}{-2.0}
    & \bvimp{70.3}{+7.4}
    & 79.1 \\
    { + [ICLR'26] SHIELD}
    & \vimp{54.5}{+5.1}
    & \vbad{8.5}{+1.0}
    & 42.5 & 4.0
    & 64.5
    & \vimp{93.5}{+1.0}
    & \vbad{50.0}{-12.9}
    & 65.2 \\
    { + [CVPR'26] PND}
    & \vimp{51.7}{+2.3}
    & \vimp{7.3}{-0.2}
    & 34.9 & 4.0
    & 71.4
    & \vimp{93.2}{+0.7}
    & \vbad{61.3}{-1.6}
    & 74.0 \\
    \rowcolor{myblue}
    \textbf{ + PatchGate (Ours)}
    & \bvimp{56.0}{+6.6}
    & \vimp{6.6}{-0.9}
    & 34.3 & 3.0
    & 74.5
    & \vimp{93.8}{+1.3}
    & \vimp{66.0}{+3.1}
    & 77.4 \\
    \midrule
    \rowcolor{gray!15}
    \multicolumn{9}{l}{\textbf{[CVPR'24] LLaVA-v1.5-13B}} \\
    \midrule
    {[CVPR'24] LLaVA-v1.5-13B}
    & 50.8 & 6.5 & 30.8 & 3.1
    & 71.5 & 96.0 & 59.5 & 73.5 \\
    { + [CVPR'24] OPERA}
    & \vbad{49.0}{-1.8}
    & \vimp{6.1}{-0.4}
    & 28.4 & 2.7
    & 73.8
    & \vbad{93.1}{-2.9}
    & \vimp{63.9}{+4.4}
    & 75.8 \\
    { + [CVPR'24] VCD}
    & \vimp{51.5}{+0.7}
    & \vbad{8.0}{+1.5}
    & 35.2 & 3.7
    & \textbf{82.9}
    & \vbad{88.0}{-8.0}
    & \bvimp{86.0}{+26.5}
    & \textbf{87.0} \\
    { + [CVPR'25] Devils-in-Mid.}
    & \vbad{46.3}{-4.5}
    & \bvimp{4.3}{-2.2}
    & \textbf{18.9} & \textbf{1.2}
    & 63.3
    & \vbad{95.3}{-0.7}
    & \vbad{46.9}{-12.6}
    & 62.9 \\
    { + [ICML'25] MARINE}
    & \vbad{50.0}{-0.8}
    & \vimp{5.0}{-1.5}
    & 23.9 & 2.3
    & 70.9
    & \bvimp{96.7}{+0.7}
    & \vbad{58.0}{-1.5}
    & 72.5 \\
    { + [ICLR'25] ProjectAway}
    & \vbad{49.5}{-1.3}
    & \vbad{8.6}{+2.1}
    & 35.8 & 4.6
    & 70.2
    & \vbad{92.3}{-3.7}
    & \vimp{60.1}{+0.6}
    & 72.8 \\
    { + [ICML'26] ILVAD}
    & \vimp{50.9}{+0.1}
    & \vimp{5.1}{-1.4}
    & 27.0 & 2.2
    & 74.2
    & \vbad{90.8}{-5.2}
    & \vimp{64.3}{+4.8}
    & 75.3 \\
    { + [ICLR'26] SHIELD}
    & \vimp{51.3}{+0.5}
    & \vbad{9.0}{+2.5}
    & 41.4 & 4.4
    & 64.8
    & \vbad{93.2}{-2.8}
    & \vbad{50.4}{-9.1}
    & 65.4 \\
    { + [CVPR'26] PND}
    & \vimp{52.1}{+1.3}
    & \vbad{7.0}{+0.5}
    & 33.2 & 3.7
    & 72.3
    & \vbad{93.0}{-3.0}
    & \vimp{62.1}{+2.6}
    & 74.5 \\
    \rowcolor{myblue}
    \textbf{ + PatchGate (Ours)}
    & \bvimp{52.8}{+2.0}
    & \vimp{5.7}{-0.8}
    & 25.7 & 2.2
    & 71.7
    & \vimp{96.6}{+0.6}
    & \vbad{59.3}{-0.2}
    & 73.5 \\
    \midrule
    \rowcolor{gray!15}
    \multicolumn{9}{l}{\textbf{[arXiv'25] Qwen2.5-VL-7B}} \\
    \midrule
    {[arXiv'25] Qwen2.5-VL-7B}
    & 58.8 & 4.1 & 27.2 & 1.7
    & 84.9 & 92.9 & 83.6 & 88.0 \\
    { + [CVPR'24] OPERA}
    & \vbad{57.2}{-1.6}
    & \vbad{4.4}{+0.3}
    & 27.9 & 1.6
    & 85.1
    & \vbad{92.4}{-0.5}
    & \vimp{84.0}{+0.4}
    & 88.0 \\
    { + [CVPR'24] VCD}
    & \vbad{53.7}{-5.1}
    & \vbad{6.3}{+2.2}
    & 34.6 & 1.8
    & 81.2
    & \vbad{83.9}{-9.0}
    & \vimp{88.7}{+5.1}
    & 86.2 \\
    { + [CVPR'25] Devils-in-Mid.}
    & \vbad{54.8}{-4.0}
    & \vbad{4.2}{+0.1}
    & 17.5 & 1.0
    & 84.9
    & \bvimp{93.7}{+0.8}
    & \vbad{82.8}{-0.8}
    & 87.9 \\
    { + [ICML'25] MARINE}
    & \bvimp{62.4}{+3.6}
    & \vbad{4.4}{+0.3}
    & 25.8 & 1.7
    & \textbf{89.0}
    & \vbad{92.1}{-0.8}
    & \vimp{91.3}{+7.7}
    & \textbf{91.7} \\
    { + [ICLR'25] ProjectAway}
    & \vbad{55.5}{-3.3}
    & \vbad{4.9}{+0.8}
    & 31.8 & 1.8
    & 84.0
    & \vbad{85.5}{-7.4}
    & \vimp{88.0}{+4.4}
    & 86.7 \\
    { + [ICML'26] ILVAD}
    & \vbad{35.9}{-22.9}
    & \bvimp{3.1}{-1.0}
    & \textbf{7.2} & \textbf{0.3}
    & 82.5
    & \vbad{91.3}{-1.6}
    & \vbad{81.4}{-2.2}
    & 86.1 \\
    { + [ICLR'26] SHIELD}
    & \vbad{53.1}{-5.7}
    & \vbad{5.8}{+1.7}
    & 33.3 & 1.2
    & 86.2
    & \vbad{90.3}{-2.6}
    & \vimp{85.9}{+2.3}
    & 88.0 \\
    { + [CVPR'26] PND}
    & \vbad{55.3}{-3.5}
    & \vbad{5.0}{+0.9}
    & 29.8 & 1.6
    & 83.4
    & \vbad{90.1}{-2.8}
    & \vimp{84.2}{+0.6}
    & 87.0 \\
    \rowcolor{myblue}
    \textbf{ + PatchGate (Ours)}
    & \vimp{60.2}{+1.4}
    & \vimp{3.6}{-0.5}
    & 21.8 & 1.3
    & 85.4
    & \vbad{84.7}{-8.2}
    & \bvimp{95.2}{+11.6}
    & 89.6 \\
    \bottomrule
    \end{tabular}
    }
\end{table}

\begin{table}[h]
    \centering
    \caption{Performance comparison on the POPE benchmark~\cite{li2023pope}. We report results across the random, popular, and adversarial splits. The best results are in \textbf{bold}, and the second-best results are \underline{underlined}.}
    \label{tab:ap_pope_splitwise}
    \resizebox{\linewidth}{!}{
    \begin{tabular}{lcccccccccccc}
    \toprule
    \multirow{2}{*}{\textbf{Method}}
    & \multicolumn{4}{c}{\textbf{Random}}
    & \multicolumn{4}{c}{\textbf{Popular}}
    & \multicolumn{4}{c}{\textbf{Adversarial}} \\
    \cmidrule(lr){2-5} \cmidrule(lr){6-9} \cmidrule(lr){10-13}
    & \textbf{Acc.} $\uparrow$
    & \textbf{P.} $\uparrow$
    & \textbf{R.} $\uparrow$
    & \textbf{F1} $\uparrow$
    & \textbf{Acc.} $\uparrow$
    & \textbf{P.} $\uparrow$
    & \textbf{R.} $\uparrow$
    & \textbf{F1} $\uparrow$
    & \textbf{Acc.} $\uparrow$
    & \textbf{P.} $\uparrow$
    & \textbf{R.} $\uparrow$
    & \textbf{F1} $\uparrow$ \\
    \midrule
    {[CVPR'24] LLaVA-v1.5-7B}
    & 83.8 & 92.4 & 73.7 & 82.0
    & 82.6 & 89.7 & 73.7 & 80.9
    & 79.7 & 83.8 & 73.7 & 78.4 \\
    { + [CVPR'24] OPERA}
    & 89.2 & 93.4 & 85.2 & 89.1
    & 86.8 & 88.0 & 85.2 & {86.6}
    & 81.2 & 78.8 & 85.2 & 81.9 \\
    { + [CVPR'24] VCD}
    & 85.4 & 92.8 & 76.8 & 84.0
    & 83.3 & 88.3 & 76.8 & 82.2
    & 80.4 & 82.8 & 76.8 & 79.7 \\
    { + [CVPR'25] Devils-in-Mid.}
    & 56.0 & 54.6 & 70.3 & 61.5
    & 56.2 & 54.8 & 70.9 & 61.8
    & 54.5 & 53.5 & 69.1 & 60.3 \\
    { + [ICML'25] MARINE}
    & 86.5 & 91.0 & 79.0 & 84.6
    & 84.5 & 89.5 & 78.0 & 83.4
    & 82.2 & \underline{87.7} & 77.0 & 82.0 \\
    { + [ICLR'25] ProjectAway}
    & 89.0 & 89.4 & \underline{89.2} & 89.3
    & 85.0 & 82.3 & \textbf{89.2} & 85.6
    & 79.0 & 74.0 & \textbf{89.2} & 80.9 \\
    { + [ICML'26] ILVAD}
    & 89.2 & 91.4 & 86.6 & 88.9
    & 86.6 & 86.6 & 86.6 & 86.6
    & 80.9 & 77.7 & 86.6 & 81.9 \\
    { + [ICLR'26] SHIELD}
    & \underline{90.4} & 91.9 & 88.6 & \underline{90.2}
    & 87.0 & 85.9 & 88.6 & \underline{87.2}
    & 81.7 & 77.9 & \underline{88.6} & 82.9 \\
    { + [CVPR'26] PND}
    & 88.6 & \textbf{96.3} & 80.3 & 87.6
    & \underline{87.2} & \textbf{93.1} & 80.3 & 86.2
    & \underline{85.1} & \textbf{88.8} & 80.3 & \underline{84.3} \\
    \rowcolor{myblue}
    \textbf{ + PatchGate (Ours)}
    & \textbf{92.5} & \underline{94.8} & \textbf{89.9} & \textbf{92.3}
    & \textbf{89.9} & \underline{90.7} & \underline{89.0} & \textbf{89.8}
    & \textbf{87.0} & 86.3 & 87.9 & \textbf{87.1} \\
    \bottomrule
    \end{tabular}
    }
\end{table}

\paragraph{Prompt and inference protocols.}
VEX receives only the image \(I\) and performs an additional image-only forward pass without textual input or output-token decoding.
Task-specific prompts are introduced only during downstream inference.

\textbf{Generative captioning.}
For AMBER generative captioning, we use the official prompt, ``Describe this image.''~\cite{wang2023amber}
For this open-ended task, VIED directly applies the ESI and EDE object-token logit updates described in Sec.~\ref{subsec:vied} and Algorithm~\ref{alg:vied} throughout greedy decoding.

\textbf{Binary QA.}
For the AMBER discriminative task, we preserve each query from the official JSON file without modifying its wording.
The queries cover object existence, attributes, and relations, such as ``Is there a cloud in this image?'', ``Is the sky sunny in this image?'', and ``Is there direct contact between the person and grass?'', respectively.
For POPE, we use ``Is there a \{object\} in the image? Please answer this question with one word.''
The unconditioned inference uses the original image and benchmark query, whereas the conditioned inference additionally receives the image-specific inventory as the fixed context ``The image contains: \{object$_1$, object$_2$, \ldots\}.'' (see Fig.~\ref{fig:worked_example_binaryqa}).
The final prediction is restricted to the ``Yes'' and ``No'' answer tokens.

\paragraph{Runs and environment.}
We report a single inference run for each method and benchmark.
PatchGate uses deterministic greedy decoding with \texttt{do\_sample=False}, without temperature, \texttt{top\_p}, \texttt{num\_beams}, or a random seed.
For stochastic baselines, we use a fixed seed of \(42\) and temperature \(1.0\).
All experiments run on one NVIDIA RTX A6000 GPU with \rev{48GB} memory using Python 3.10.12, PyTorch 2.1.2, and Transformers 4.45.2.

\paragraph{Code archive.}
The submitted code archive includes the complete PatchGate inference procedures for generative captioning and binary QA, together with configuration files and a \texttt{README.md} containing setup and execution instructions.

\FloatBarrier
\section{Additional Quantitative Results} \label{app:quantitative_results}

\subsection{Additional AMBER Results} \label{app:additional_amber}

\paragraph{Overall performance across backbones.}
Tab.~\ref{tab:ap_amber} reports the full AMBER~\cite{wang2023amber} results on the main LLaVA-v1.5-7B backbone~\cite{liu2024llavaimproved}, the larger LLaVA-v1.5-13B, and Qwen2.5-VL-7B~\cite{Qwen2.5-VL} from a different model family.
On LLaVA-v1.5-7B, PatchGate improves Cover from \(49.4\) to \(56.0\) (\(+13.4\%\)) while reducing CHAIR from \(7.5\) to \(6.6\) (\(-12.0\%\)).
The same trend appears on LLaVA-v1.5-13B and Qwen2.5-VL-7B, demonstrating consistent mitigation of omission and hallucination across model scales and families.
AMBER discriminative QA is less directly aligned with PatchGate's primary objective: its object-existence subset evaluates only absent-object rejection, while its attribute and relation subsets require evidence beyond object presence. Despite this mismatch, PatchGate remains competitive, ranking third in Accuracy, Precision, Recall, and F1 on LLaVA-v1.5-7B. This suggests that PatchGate's image-specific object evidence remains useful for binary QA, even when the task extends beyond its primary object-verbalization objective.

\paragraph{Balanced improvement without metric trade-offs.}
Existing methods often improve one aspect of object reliability at the expense of its complementary objective.
For example, ILVAD~\cite{Xie2026ilvad} substantially reduces CHAIR but also decreases Cover, while Devils-in-Mid.~\cite{jiang2025devils} and MARINE~\cite{zhao2025marine} improve Precision at the cost of Recall.
Conversely, SHIELD~\cite{huang2026shield} increases Cover but worsens CHAIR and substantially reduces Recall.
In contrast, PatchGate is the only evaluated method on LLaVA-v1.5-7B that simultaneously improves Cover, Precision, and Recall while reducing CHAIR relative to the frozen backbone.
This demonstrates balanced gains across generative object coverage, mention-level hallucination, and discriminative QA, while the POPE results provide direct evidence of bidirectional correction for present and absent objects.

\paragraph{Aggregate improvement across primary metrics.}
To assess balanced improvement without overemphasizing any single metric, we conduct a simple post-hoc analysis that equally sums the percentage-point changes in Cover, CHAIR, Precision, and Recall, counting a reduction in CHAIR as positive (see the red and green highlights in Tab.~\ref{tab:ap_amber}).
PatchGate achieves the largest aggregate gain of \(6.6 + 0.9 + 1.3 + 3.1 = 11.9\), while the second-highest gain is obtained by OPERA~\cite{huang2024opera} with \(-0.6 + 0.9 + 0.2 + 7.0 = 7.5\).
This comparison further highlights that PatchGate improves the four primary measures jointly without trading off one objective against another.

\subsection{Additional POPE Results} \label{app:additional_pope}
The main paper reports POPE~\cite{li2023pope} results averaged over the random, popular, and adversarial splits, and we further examine the full split-wise results (see Tab.~\ref{tab:ap_pope_splitwise}).
Unlike AMBER discriminative QA, POPE directly matches PatchGate's bidirectional object-level objective by including both present- and absent-object questions.
\paragraph{Split-wise gains over the baseline.}
Across the random, popular, and adversarial splits, respectively, PatchGate improves Precision by \(2.6\%\), \(1.1\%\), and \(3.0\%\), and Recall by \(22.0\%\), \(20.8\%\), and \(19.3\%\) over LLaVA-v1.5-7B, indicating fewer hallucinated ``Yes'' answers and fewer incorrect ``No'' answers for visible objects, respectively.
These consistent gains yield the best Accuracy and F1 in every split.

\paragraph{Precision-recall trade-off of competing methods.}
The metric-specific best results achieved by competing methods exhibit a pronounced Precision-Recall trade-off.
PND~\cite{jiang2026pnd} achieves the highest Precision in all three splits, but its Recall remains \(7.6\)--\(9.6\) points below PatchGate.
Conversely, ProjectAway~\cite{jiang2025projectaway} achieves the highest Recall on the popular and adversarial splits, but its Precision falls \(8.4\) and \(12.3\) points below PatchGate, respectively.
Overall, PatchGate ranks first in \(7\), second in \(3\), and third in the remaining \(2\) of the \(12\) split-wise metrics, demonstrating the strongest balance between suppressing hallucinated affirmative answers and correctly affirming visible objects.

\FloatBarrier
\section{Additional Qualitative Results} \label{app:qualitative_results}
Beyond the examples in Fig.~\ref{fig:qualitative_results} of the main paper, we provide additional qualitative comparisons with representative training-free hallucination-mitigation methods on AMBER~\cite{wang2023amber} and POPE~\cite{li2023pope}.

\begin{figure}[p]
    \centering
    \includegraphics[width=0.8\linewidth]{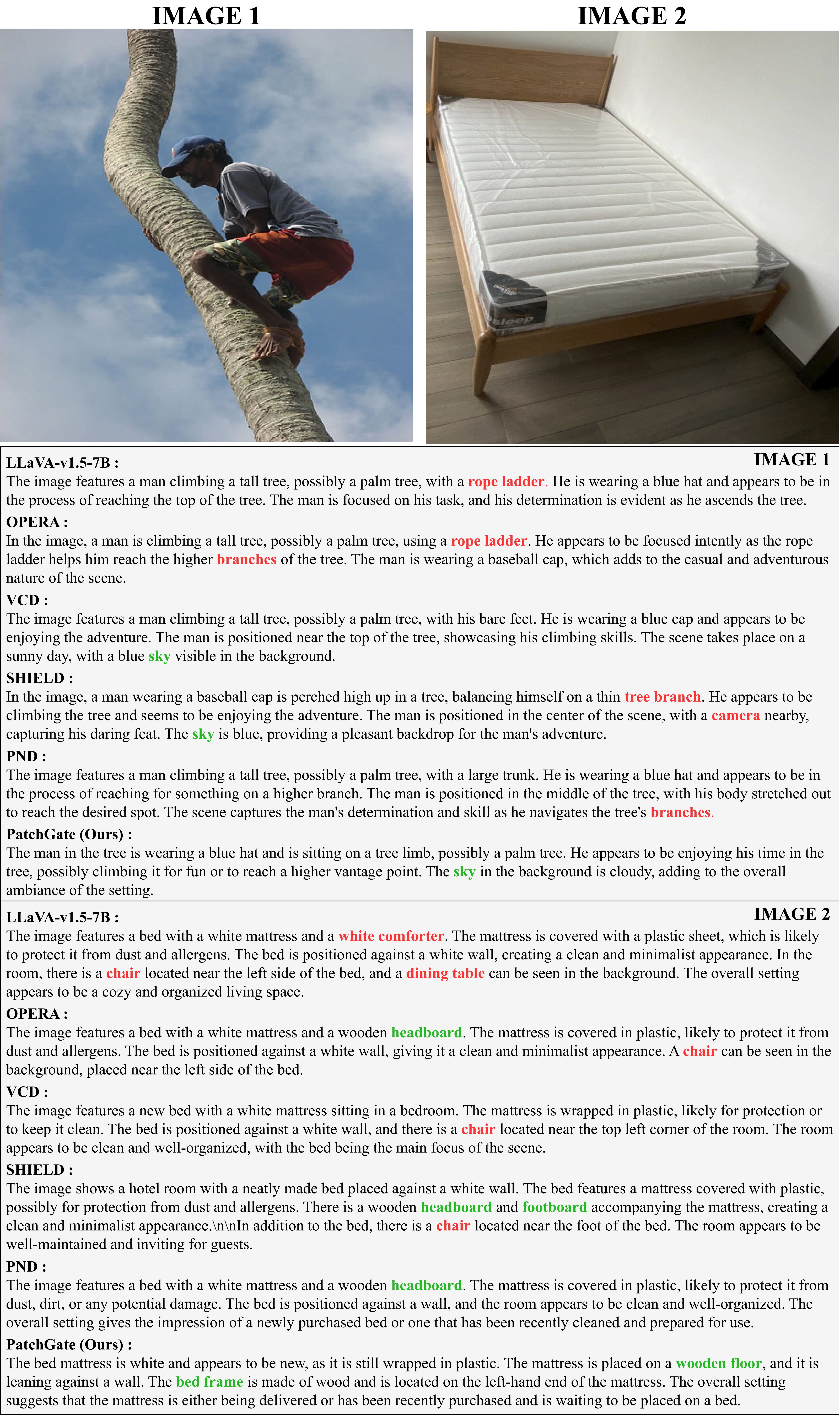}
    \caption{\textbf{Additional qualitative results on AMBER generative task.}
    PatchGate suppresses unsupported object mentions (red) while recovering omitted visible objects(green), generating precise and complete captions.}
    \label{fig:app_qual_amber}
\end{figure}

\paragraph{Captioning examples.}
% Fig.~\ref{fig:app_qual_amber} presents additional AMBER captioning examples.
% PatchGate removes unsupported object mentions while recovering visible content omitted by the frozen backbone, demonstrating balanced mitigation of hallucination and omission.
Fig.~\ref{fig:app_qual_amber} presents additional AMBER captioning examples.
Suppression-oriented baselines remove some unsupported mentions but still introduce new ones, such as \emph{rope ladder}, \emph{branches}, and \emph{camera} in the first image and \emph{white comforter} and \emph{dining table} in the second, and they provide no mechanism to restore omitted content.
PatchGate instead removes these unsupported object mentions while recovering visible content omitted by the frozen backbone, such as \emph{sky} in the first image and \emph{wooden floor} and \emph{bed frame} in the second, demonstrating balanced mitigation of hallucination and omission.

\begin{figure}[p]
    \centering
    \includegraphics[width=0.8\linewidth]{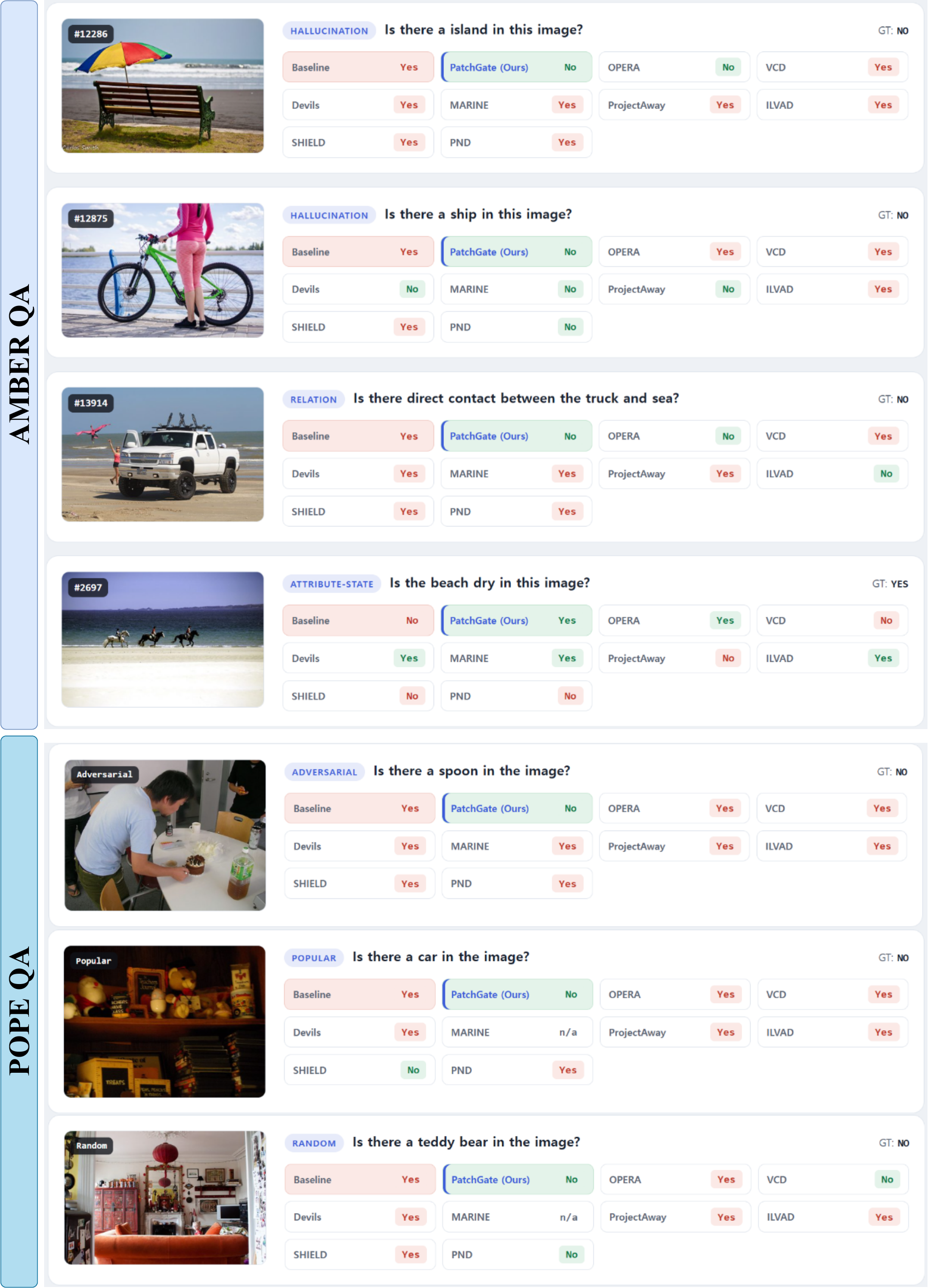}
    \caption{\textbf{Additional qualitative results on AMBER and POPE QA.} Correct and incorrect answers are shown in green and red, respectively. PatchGate corrects both hallucinated ``Yes'' answers and omitted ``No'' answers across diverse question types.}
    \label{fig:app_qual_qa}
\end{figure}

\paragraph{Binary QA examples.}
% Fig.~\ref{fig:app_qual_qa} presents AMBER questions covering object existence and attributes, together with POPE examples.
% PatchGate corrects both hallucinated ``Yes'' answers for absent objects and incorrect ``No'' answers for visible objects.
Fig.~\ref{fig:app_qual_qa} presents AMBER questions covering object existence, attributes, and relations, together with POPE examples from the random, popular, and adversarial splits.
Most baselines exhibit a ``Yes'' bias on absent-object questions, answering affirmatively for objects such as \emph{island}, \emph{ship}, \emph{spoon}, and \emph{teddy bear}.
PatchGate corrects both directions: it flips these hallucinated ``Yes'' answers to ``No'', while also fixing incorrect ``No'' answers on questions about visible content, rather than trading one error type for the other.

\paragraph{Failure patterns of suppression-only baselines.}
A closer look at Figs.~\ref{fig:app_qual_amber} and~\ref{fig:app_qual_qa} reveals a consistent pattern behind these corrections.
Suppression-oriented methods reduce hallucinated mentions largely by making captions shorter and less specific, which leaves omissions untouched and can even drop correctly mentioned objects.
PatchGate instead edits logits in both directions against the same pre-generation inventory, so suppression and recovery are coupled rather than traded off: captions remain similarly detailed while their object mentions align more closely with the image.
This qualitative behavior mirrors the quantitative trends in Appendix~\ref{app:quantitative_results}, where PatchGate improves Cover and Recall together with CHAIR and Precision.

\FloatBarrier
\section{Details of Ablation Study} \label{app:ablation_study}
This section examines how the main design choices of PatchGate affect the balance between omission recovery and hallucination suppression. Unless otherwise specified, all analyses use the AMBER~\cite{wang2023amber} generative task with the LLaVA-v1.5-7B backbone~\cite{liu2024llavaimproved}.

\begin{table}[h]
    \centering
    \caption{Effect of VIED inclusion and exclusion strengths.}
    \label{tab:app_vied_strength_sensitivity}
    \adjustbox{max width=\linewidth}{
    \begin{tabular}{cc|cccc}
    \toprule
    \multicolumn{2}{c|}{\textbf{VIED Hyperparameters}}
    & \multicolumn{4}{c}{\textbf{Metrics}} \\
    \cmidrule(lr){1-2} \cmidrule(lr){3-6}
    \(\boldsymbol{\alpha_{\mathrm{esi}}}\)
    & \(\boldsymbol{\gamma_{\mathrm{ede}}}\)
    & \textbf{Cover} $\uparrow$
    & \textbf{CHAIR} $\downarrow$
    & \textbf{Hal} $\downarrow$
    & \textbf{Cog} $\downarrow$ \\
    \midrule
    \rowcolor{gray!15}
    \multicolumn{6}{c}{\(\gamma_{\mathrm{ede}}=0.5\) fixed} \\
    \midrule
    2  & 0.5 & 51.6 & \textbf{6.1} & \textbf{30.7} & 3.2 \\
    4  & 0.5 & 52.8 & 6.3 & 31.5 & 3.1 \\
    \rowcolor{myblue}
    \textbf{8} & \textbf{0.5} & 56.0 & 6.6 & 34.3 & \textbf{3.0} \\
    16 & 0.5 & \textbf{58.8} & 6.8 & 36.4 & 3.1 \\
    \midrule
    \rowcolor{gray!15}
    \multicolumn{6}{c}{\(\alpha_{\mathrm{esi}}=8\) fixed} \\
    \midrule
    8 & 0.25 & \textbf{56.3} & 6.8 & 36.4 & 3.0 \\
    \rowcolor{myblue}
    \textbf{8} & \textbf{0.5} & 56.0 & 6.6 & 34.3 & 3.0 \\
    8 & 1.0 & 54.6 & \textbf{6.5} & \textbf{32.4} & 2.8 \\
    8 & 2.0 & 52.5 & 7.0 & 33.1 & \textbf{2.7} \\
    \bottomrule
    \end{tabular}
    }
\end{table}

\paragraph{Effect of VIED hyperparameters.}
Varying the inclusion strength reveals a clear trade-off between recovering visible objects and introducing additional object mentions (see Tab.~\ref{tab:app_vied_strength_sensitivity}).
With \(\gamma_{\mathrm{ede}}=0.5\) fixed, increasing \(\alpha_{\mathrm{esi}}\) from \(2\) to \(16\) raises Cover from \(51.6\) to \(58.8\), while CHAIR and Hal gradually increase.
We therefore select \(\alpha_{\mathrm{esi}}=8\), which provides a substantial coverage gain without the larger hallucination increase observed at \(\alpha_{\mathrm{esi}}=16\).

A moderate exclusion strength improves hallucination suppression while retaining most of the recovered content.
Increasing \(\gamma_{\mathrm{ede}}\) from \(0.25\) to \(1.0\) reduces CHAIR from \(6.8\) to \(6.5\) and Hal from \(36.4\) to \(32.4\), but also lowers Cover from \(56.3\) to \(54.6\).
At \(\gamma_{\mathrm{ede}}=2.0\), Cover decreases further and CHAIR rises to \(7.0\), indicating that excessive suppression can disrupt otherwise valid generation.
We use \(\gamma_{\mathrm{ede}}=0.5\) as a conservative operating point that preserves the coverage improvement while maintaining effective hallucination control.

\begin{table}[h]
    \centering
    \caption{Sensitivity to the QA guidance strength \(\lambda_{\mathrm{QA}}\) on AMBER discriminative QA.}
    \label{tab:qa_lambda}
    \adjustbox{max width=\linewidth}{
    \begin{tabular}{lcccc}
    \toprule
    \(\boldsymbol{\lambda_{\mathrm{QA}}}\) & \textbf{Acc.} $\uparrow$ & \textbf{P.} $\uparrow$ & \textbf{R.} $\uparrow$ & \textbf{F1} $\uparrow$ \\
    \midrule
    0.0 (base)
    & 72.0 & 92.5 & 62.9 & 74.9 \\
    0.3
    & 73.2 & 92.9 & 64.6 & 76.2 \\
    0.5
    & 73.4 & 93.5 & 65.7 & 77.2 \\
    \rowcolor{myblue}
    \textbf{0.7}
    & \textbf{74.5} & \textbf{93.8} & 66.0 & \textbf{77.4} \\
    0.9
    & 74.4 & 93.2 & 66.2 & \textbf{77.4} \\
    1.0
    & 74.4 & 93.1 & \textbf{66.3} & \textbf{77.4} \\
    \bottomrule
    \end{tabular}
    }
\end{table}

\paragraph{Binary-QA guidance strength.}
Binary-QA inference combines the conditioned and unconditioned logits using the QA guidance strength \(\lambda_{\mathrm{QA}}\), as defined in Eq.~\ref{eq:qa_guidance}.
Increasing \(\lambda_{\mathrm{QA}}\) from \(0\) to \(0.7\) improves Accuracy from \(72.0\) to \(74.5\), Precision from \(92.5\) to \(93.8\), Recall from \(62.9\) to \(66.0\), and F1 from \(74.9\) to \(77.4\) on AMBER discriminative QA (see Tab.~\ref{tab:qa_lambda}).
Larger values provide only a marginal Recall gain, while Accuracy and Precision slightly decrease and F1 remains unchanged.
We therefore use \(\lambda_{\mathrm{QA}}=0.7\), which provides the strongest overall balance between the original answer distribution and the inventory-conditioned signal.

\begin{table}[ht]
    \centering
    \caption{Linguistic quality of PatchGate on AMBER.}
    \label{tab:app_fluency}
    \adjustbox{max width=\linewidth}{
    \begin{tabular}{lccc}
    \toprule
    \textbf{Method}
    & \textbf{PPL} $\downarrow$
    & \textbf{Rep-4} $\downarrow$
    & \textbf{Avg. Len.} \\
    \midrule
    LLaVA-v1.5-7B
    & \textbf{4.21} & \textbf{0.018} & 52.3 \\
    \rowcolor{myblue}
    \textbf{ + PatchGate}
    & 4.35 & 0.020 & 58.1 \\
    \midrule
    LLaVA-v1.5-13B
    & \textbf{3.98} & \textbf{0.016} & 54.0 \\
    \rowcolor{myblue}
    \textbf{ + PatchGate}
    & 4.10 & 0.018 & 59.2 \\
    \midrule
    Qwen2.5-VL-7B
    & \textbf{3.62} & \textbf{0.012} & 61.5 \\
    \rowcolor{myblue}
    \textbf{ + PatchGate}
    & 3.70 & 0.013 & 64.8 \\
    \midrule
    InstructBLIP-7B
    & \textbf{5.80} & \textbf{0.041} & 38.2 \\
    \rowcolor{myblue}
    \textbf{ + PatchGate}
    & 5.95 & 0.043 & 42.0 \\
    \bottomrule
    \end{tabular}
    }
\end{table}

\paragraph{Linguistic quality.}
PatchGate operates consistently across different model scales and VLM families in the main experiments (see Tab.~\ref{tab:ab_robustness_vlm_backbone}).
Because it directly adjusts object-token logits, we further evaluate sentence-level fluency using perplexity (PPL), 4-gram repetition (Rep-4), and average response length across these backbones (see Tab.~\ref{tab:app_fluency}).
Lower PPL and Rep-4 indicate more probable and less repetitive generation, respectively.
Across all evaluated backbones, PatchGate produces longer responses with modest increases in PPL and Rep-4.
Although these metrics do not identify the precise linguistic source of the changes, their small magnitude indicates that the gains in object reliability do not substantially degrade overall fluency.

\begin{table}[ht]
    \centering
    \caption{Computational cost comparison.}
    \label{tab:app_inference_cost}
    \adjustbox{max width=\linewidth}{
    \begin{tabular}{lccc}
    \toprule
    \textbf{Method}
    & \shortstack{\textbf{External}\\\textbf{Module}}
    & \shortstack{\textbf{Peak}\\\textbf{VRAM} $\downarrow$}
    & \shortstack{\textbf{Inference}\\\textbf{Time} $\downarrow$} \\
    \midrule
    {[CVPR'24] LLaVA-v1.5-7B}
    & No & \textbf{14.6 GB} & \textbf{2.92 s} \\
    { + [CVPR'24] OPERA}
    & No & 16.8 GB & 24.50 s \\
    { + [CVPR'24] VCD}
    & No & 15.2 GB & 5.10 s \\
    { + [CVPR'25] Devils-in-Mid.}
    & No & 15.0 GB & 3.05 s \\
    { + [ICML'25] MARINE}
    & Yes & 17.5 GB & 3.60 s \\
    { + [ICLR'25] ProjectAway}
    & No & 15.1 GB & 3.20 s \\
    { + [ICML'26] ILVAD}
    & No & 15.3 GB & 4.20 s \\
    { + [ICLR'26] SHIELD}
    & No & 15.4 GB & 5.80 s \\
    { + [CVPR'26] PND}
    & No & 17.2 GB & 8.84 s \\
    \rowcolor{myblue}
    \textbf{+ PatchGate (Ours)}
    & No & 14.9 GB & 3.30 s \\
    \bottomrule
    \end{tabular}
    }
\end{table}

\begin{figure}[!t]
    \centering
    \includegraphics[width=0.85\columnwidth]{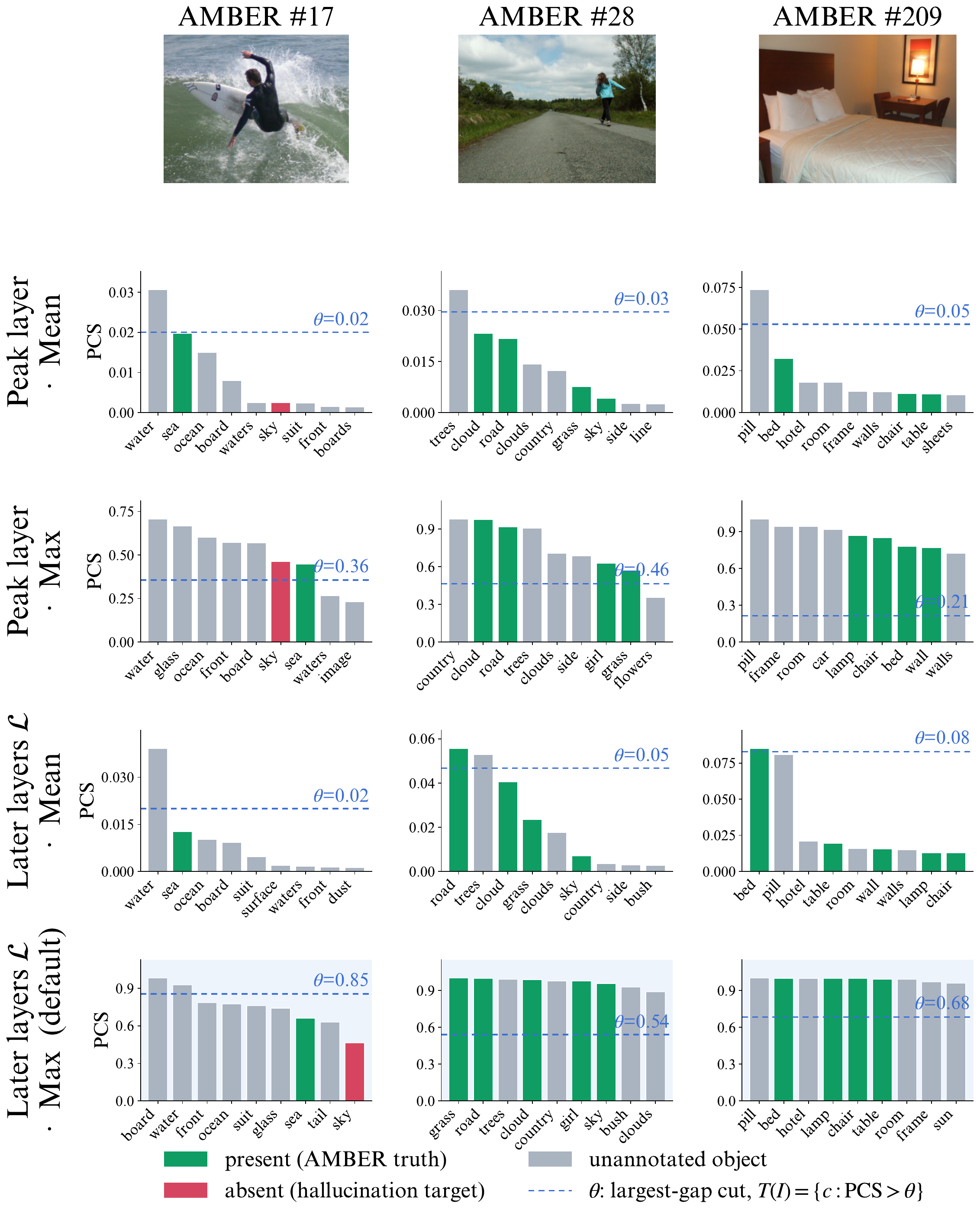}
    \caption{\textbf{Evidence aggregation and adaptive object selection.}
    Mean aggregation retains only a few high-confidence objects, whereas max-softmax aggregation over the selected late-layer range more clearly separates supported objects from the low-evidence tail.}
    \label{fig:TTD}
\end{figure}

\begin{table}[t]
    \centering
    \caption{Effect of the VEX layer source and evidence aggregation on AMBER.}
    \label{tab:ap_vex_aggregation}
    \adjustbox{max width=\linewidth}{
    \begin{tabular}{llcccccccc}
    \toprule
    \multirow{2}{*}{\textbf{Layer Source}}
    & \multirow{2}{*}{\textbf{Aggregation}}
    & \multicolumn{4}{c}{\textbf{Tag-level}}
    & \multicolumn{4}{c}{\textbf{Caption-level}} \\
    \cmidrule(lr){3-6} \cmidrule(lr){7-10}
    &
    & \textbf{P.} $\uparrow$
    & \textbf{R.} $\uparrow$
    & \textbf{F1} $\uparrow$
    & \(\mathbf{|T|}\)
    & \textbf{Cover} $\uparrow$
    & \textbf{CHAIR} $\downarrow$
    & \textbf{Hal} $\downarrow$
    & \textbf{Cog} $\downarrow$ \\
    \midrule
    single peak late layer
    & Mean
    & \textbf{97.7} & 9.0 & 16.4 & 1.0
    & 32.1 & 11.2 & \textbf{30.0} & 2.4 \\
    single peak late layer
    & Max
    & 89.5 & 31.9 & 47.0 & 6.3
    & 49.3 & 7.1 & 32.5 & 2.9 \\
    Later layers \(\mathcal{L}\)
    & Mean
    & 97.6 & 10.2 & 18.4 & 1.1
    & 32.8 & 11.2 & 30.4 & \textbf{2.3} \\
    \rowcolor{myblue}
    Later layers \(\mathcal{L}\)
    & Max
    & 84.8 & \textbf{49.2} & \textbf{62.3} & 12.9
    & \textbf{56.0} & \textbf{6.6} & 34.3 & 3.0 \\
    \bottomrule
    \end{tabular}
    }
\end{table}

\begin{table}[t]
    \centering
    \caption{Effect of VEX layer range.}
    \label{tab:ab_vex_layer_range}
    \adjustbox{max width=\linewidth}{
    \begin{tabular}{lcccccccc}
    \toprule
    \multirow{2}{*}{\textbf{Layer Range}}
    & \multicolumn{4}{c}{\textbf{Tag-level}}
    & \multicolumn{4}{c}{\textbf{Caption-level}} \\
    \cmidrule(lr){2-5} \cmidrule(lr){6-9}
    & \textbf{P.} $\uparrow$
    & \textbf{R.} $\uparrow$
    & \textbf{F1} $\uparrow$
    & \(\mathbf{|T|}\)
    & \textbf{Cover} $\uparrow$
    & \textbf{CHAIR} $\downarrow$
    & \textbf{Hal} $\downarrow$
    & \textbf{Cog} $\downarrow$ \\
    \midrule
    All layers \((0\%-100\%)\)
    & 83.4 & \textbf{49.6} & 62.2 & 13.5
    & 55.8 & 7.0 & 35.1 & 3.0 \\
    Later \(67\%\) \((33\%-100\%)\)
    & 83.4 & 49.5 & 62.1 & 13.4
    & 55.7 & 6.9 & 35.0 & \textbf{2.9} \\
    \rowcolor{myblue}
    Later \(33\%\) \((67\%-100\%)\)
    & \textbf{84.8} & {49.2} & \textbf{62.3} & {12.9}
    & \textbf{56.0} & \textbf{6.6} & \textbf{34.3} & 3.0 \\
    \bottomrule
    \end{tabular}
    }
\end{table}

\paragraph{Computational efficiency.}
Extracting visual evidence directly from the frozen target VLM allows PatchGate to avoid any external detection or tagging module while introducing limited computational overhead (see Tab.~\ref{tab:app_inference_cost}).
Relative to LLaVA-v1.5-7B, it increases peak VRAM by only \(0.3\) GB (\(2.1\%\)) and inference time by \(0.38\) seconds (\(13.0\%\)).
Its peak memory of \(14.9\) GB is the lowest among the evaluated hallucination-mitigation methods, while its \(3.30\)-second inference time is shorter than those of most competing approaches.
These results indicate that PatchGate improves object reliability with modest memory and runtime overhead, without introducing an external visual model.

\begin{table}[t]
    \centering
    \caption{Comparison of internal (VEX; Sec.~\ref{subsec:vex}) and external object inventories.}
    \label{tab:ap_ab_vex_effect}
    \resizebox{\linewidth}{!}{
    \begin{tabular}{lccccccccccc}
    \toprule
    \multirow{2}{*}{\textbf{Method}}
    & \multicolumn{4}{c}{\textbf{Tag-level}}
    & \multicolumn{4}{c}{\textbf{Caption-level}}
    & \multicolumn{3}{c}{\textbf{Computational Cost}} \\
    \cmidrule(lr){2-5} \cmidrule(lr){6-9} \cmidrule(lr){10-12}
    & \textbf{P.} $\uparrow$
    & \textbf{R.} $\uparrow$
    & \textbf{F1} $\uparrow$
    & \(\mathbf{|T|}\)
    & \textbf{Cover} $\uparrow$
    & \textbf{CHAIR} $\downarrow$
    & \textbf{Hal} $\downarrow$
    & \textbf{Cog} $\downarrow$
    & \textbf{Param.}
    & \textbf{Peak VRAM}
    & \textbf{Time} \\
    \midrule
    LLaVA-v1.5-7B
    & -- & -- & -- & --
    & 49.4 & 7.5 & \textbf{31.4} & 3.6 
    & \textbf{7.0 B} & \textbf{14.6 GB} & \textbf{2.92 s} \\
    { + GroundingDINO}
    & 91.1 & 50.4 & 64.9 & 5.2
    & 51.3 & 7.3 & 33.3 & 3.6
    & 7.2 B & 16.2 GB & 3.26 s \\
    { + RAM++ (default)}
    & \textbf{97.0} & \textbf{59.6} & \textbf{71.1} & 11.6
    & 50.2 & 7.8 & 36.8 & 4.2
    & 7.3 B & 16.9 GB & 3.17 s \\
    { + RAM++ (top-\(10\))}
    & 95.8 & 53.0 & 68.3 & 10.0
    & 50.1 & 8.2 & 37.2 & 4.3
    & 7.3 B & 16.9 GB & 3.17 s \\
    \rowcolor{myblue}
    \textbf{ + VEX (Ours; Sec.~\ref{subsec:vex})}
    & 84.8
    & 49.2
    & 62.3
    & 12.9
    & \textbf{56.0}
    & \textbf{6.6}
    & 34.3
    & \textbf{3.0}
    & \textbf{7.0 B}
    & 14.9 GB
    & 3.30 s \\
    \bottomrule
    \end{tabular}
    }
\end{table}

\paragraph{Evidence aggregation and adaptive object selection.}
The effectiveness of the largest-gap cutoff depends on whether the evidence scores preserve localized object responses while separating supported objects from the low-evidence tail (see Fig.~\ref{fig:TTD} and Tab.~\ref{tab:ap_vex_aggregation}).
Mean aggregation dilutes localized object responses across the averaged dimensions, causing the largest gap to appear near the beginning of the ranked list.
It therefore retains only about one object per image on average, yielding high tag Precision but very low Recall.
In other words, the few retained objects are generally reliable, but many visible objects are omitted from the inventory, resulting in low Cover and high CHAIR.

Max aggregation instead preserves the strongest patch-layer response for each object and produces a clearer separation between supported objects and the low-evidence tail.
Its benefit is largest when evidence is collected across the selected later layers, since different objects can exhibit their strongest vocabulary-space responses at different depths.
Compared with single-layer max aggregation, later-layer max aggregation increases tag Recall from \(31.9\) to \(49.2\), tag F1 from \(47.0\) to \(62.3\), and Cover from \(49.3\) to \(56.0\), while reducing CHAIR from \(7.1\) to \(6.6\).
We therefore apply the largest-gap cutoff to max-softmax scores aggregated over \(\mathcal{L}=\{22,\ldots,32\}\).

\paragraph{Effect of VEX layer range.}
With max aggregation fixed, restricting evidence extraction to the later \(33\%\) of the language model produces the most useful inventory for downstream decoding (see Tab.~\ref{tab:ab_vex_layer_range}).
Compared with using all layers, this setting changes tag Recall only slightly from \(49.6\) to \(49.2\), while increasing tag Precision and F1 and reducing the average inventory size.
This pattern is consistent with object evidence becoming more directly aligned with vocabulary-space predictions in later representations, whereas including earlier layers introduces responses that are less useful for output-space intervention.
The later \(33\%\) consequently achieves the best Cover and CHAIR, and we use \(\mathcal{L}=\{22,\ldots,32\}\) as the default layer range.

\paragraph{Internal and external object inventories.}
Replacing VEX with GroundingDINO~\cite{liu2024groundingdinotiny} or RAM++~\cite{huang2025ram++} preserves usable caption-level performance, indicating that PatchGate can operate with different inventory sources (see Tab.~\ref{tab:ap_ab_vex_effect}).
Despite using no external visual prior, VEX also remains competitive at the tag level, achieving an F1 of \(62.3\), only \(2.6\) points below GroundingDINO and \(6.0\)--\(8.8\) points below the RAM++ variants.
More importantly, the higher tag-level F1 of the external models does not translate into stronger downstream captioning.
VEX improves Cover to \(56.0\), compared with \(50.1\)--\(51.3\) for the external inventories, while reducing CHAIR to \(6.6\), compared with \(7.3\)--\(8.2\).

Several interface differences help characterize this result.
First, not every external tag maps one-to-one to the editable single-token object vocabulary of the target VLM, so some detected concepts cannot be directly associated with an output-token logit.
Second, the retained external tags primarily provide positive inventory membership, whereas PCS supplies a continuous evidence score over the target VLM's editable object vocabulary and can therefore support both inclusion and exclusion.
Consistent with this distinction, PCS provides stronger diagnostic separation of hallucinated and omitted objects than the thresholded RAM++ tag signal (see Tab.~\ref{tab:diag_external}).
External inventories also require additional model parameters and peak memory: GroundingDINO and RAM++ use \(16.2\) GB and \(16.9\) GB, respectively, compared with \(14.9\) GB for VEX.
Together, these results indicate that external taggers remain viable inventory sources, but higher tag-level accuracy alone does not guarantee a more effective or efficient decoding intervention.

\paragraph{Mismatch diagnosis using inventory signals.}
We further compare the signals exposed by VEX (Ours; Sec.~\ref{subsec:vex}) and RAM++ as diagnostics of the evidence--verbalization mismatch, using the same object instances as Tab.~\ref{tab:ab_mismatch}.
Hallucination is evaluated over objects mentioned in the baseline caption, whereas omission is evaluated over unmentioned objects.
RAM++ first produces per-class sigmoid scores over its \(4{,}585\)-word vocabulary and retains only the classes whose scores exceed learned class-specific thresholds.
Its resulting inventory therefore provides a binary tag decision for each vocabulary item, whereas PCS provides a continuous prompt-free evidence score within the target VLM.
Using these deployed signals, PCS achieves higher AUROC and AUPRC for both hallucination and omission, with the largest difference observed for omission (\(0.890\) vs.\ \(0.616\) AUROC; see Tab.~\ref{tab:diag_external}).
This gap suggests that graded target-model evidence better preserves the information needed to identify omitted visible objects, whereas thresholded external tags may discard weak but informative evidence and cannot cover objects outside the external vocabulary.

\begin{table}[t]
    \centering
    \caption{Diagnostic performance of the thresholded RAM++ tag decision and continuous PCS for evidence--verbalization mismatch on the same object instances as Tab.~\ref{tab:ab_mismatch}.}
    \label{tab:diag_external}
    \adjustbox{max width=\linewidth}{
    \begin{tabular}{lcccc}
    \toprule
    \multirow{2}{*}{\textbf{Diagnostic signal}}
    & \multicolumn{2}{c}{\textbf{Hallucination}}
    & \multicolumn{2}{c}{\textbf{Omission}} \\
    \cmidrule(lr){2-3}\cmidrule(lr){4-5}
    & AUROC $\uparrow$ & AUPRC $\uparrow$ & AUROC $\uparrow$ & AUPRC $\uparrow$ \\
    \midrule
    Random                                & 0.500 & 0.163 & 0.500 & \rev{0.233} \\
    RAM++
    & 0.824 & 0.379 & 0.616 & 0.392 \\
    \rowcolor{myblue}
    \textbf{PCS (Ours)}
    & \textbf{0.883} & \textbf{0.707}
    & \textbf{0.890} & \textbf{0.705} \\
    \bottomrule
    \end{tabular}
    }
\end{table}

\FloatBarrier
\section{Limitations and Future Work}
\label{app:limitations_future}
PatchGate currently represents visual evidence primarily through individual object tokens, so its largest gains appear in object-centric captioning and existence QA.
Attribute and relation questions require structured evidence that is not explicitly captured by the current inventory.
Future work will extend the inventory to object--attribute pairs and object--relation--object triplets, together with corresponding evidence-aware decoding strategies.

The predefined single-token vocabulary may also limit coverage of open-vocabulary or multi-token concepts, motivating improved evidence calibration and integration with external visual signals.
In addition, directly promoting object tokens can slightly affect local phrasing, perplexity, and repetition; fluency-aware constraints may reduce this effect.
Finally, because Hal and Cog can penalize visually valid but unannotated objects, more exhaustive annotations and human verification would provide a more reliable evaluation.

% ---- Bibliography ----
\renewcommand{\bibsection}{\section*{References}}
\bibliographystyle{arxiv2608/splncs04}
\bibliography{references}

\end{document}